\documentclass[conference]{IEEEtran}
\usepackage{graphicx}
\usepackage{amssymb}
\usepackage{amsmath}
\usepackage{color,colortbl}
\usepackage{xcolor}
\usepackage{caption}
\usepackage{url}
\usepackage[ruled,boxed,commentsnumbered]{algorithm2e}
\usepackage{verbatim}
\usepackage{graphicx}
\usepackage{amsmath}
\usepackage{verbatim}
\usepackage{tikz}
\usepackage{todonotes}
\usepackage{pgfplots}
\usepackage{algorithmicx}
\usepackage{comment}
\usepackage{diagbox}
\usepackage{booktabs}
\usepackage{bbding}
\usepackage{tikz}
\usepackage{calc}
\IEEEoverridecommandlockouts
\newtheorem{definition}{Definition}
\usepackage[colorlinks,
            linkcolor=blue,       
            anchorcolor=blue,  
            citecolor=blue,        
            ]{hyperref}
\usepackage{float}
\usepackage{multirow}
\usetikzlibrary{arrows,intersections,datavisualization}

\newtheorem{remark}{Remark}
\makeatletter
\newcommand{\linebreakand}{%
  \end{@IEEEauthorhalign}
  \hfill\mbox{}\par
  \mbox{}\hfill\begin{@IEEEauthorhalign}
}
\makeatother

\begin{document}

\title{Discovering  Representative Attribute-stars\\ via Minimum Description Length\vspace{-10pt}}

\author{
    \IEEEauthorblockN{Jiahong Liu$^{1}$\textsuperscript{$\dagger$}\thanks{\noindent\textsuperscript{$\dagger$}Work mainly done during an internship at Huawei Noah's Ark Lab.}, Min Zhou$^{2}$\textsuperscript{\textsection}, Philippe Fournier-Viger$^{3}$\textsuperscript{\textsection}, Menglin Yang$^{4}$, Lujia Pan$^{2}$, Mourad Nouioua$^{5}$}
    \IEEEauthorblockA{$^1$Harbin Institute of Technology, Shenzhen, China}
    \IEEEauthorblockA{$^2$Huawei Noah's Ark Lab, Shenzhen, China}
    \IEEEauthorblockA{$^3$Shenzhen University, Shenzhen, China}
    \IEEEauthorblockA{$^4$The Chinese University of Hong Kong, Hong Kong SAR, China}
    \IEEEauthorblockA{$^5$University of Bordj Bou Arreridj, Bordj Bou Arreridj, Algeria}
    \IEEEauthorblockA{\{jiahong.liu21, mouradnouioua\}@gmail.com, \{zhoumin27, panlujia\}@huawei.com, philfv@szu.edu.cn, mlyang@cse.cuhk.edu.hk}
}

\maketitle
\begingroup\renewcommand\thefootnote{\textsection}
\footnotetext{Corresponding Author}
\endgroup



\begin{abstract}Graphs are a popular data type found in many domains. Numerous techniques have been proposed to find interesting patterns in graphs to help understand the data and support decision-making.  However, there are generally two limitations that hinder their practical use: (1) they have multiple parameters that are hard to set but greatly influence results, (2) and they generally focus on identifying complex subgraphs while ignoring relationships between attributes of nodes.Graphs are a popular data type found in many domains. Numerous techniques have been proposed to find interesting patterns in graphs to help understand the data and support decision-making.  However, there are generally two limitations that hinder their practical use: (1) they have multiple parameters that are hard to set but greatly influence results, (2) and they generally focus on identifying complex subgraphs while ignoring relationships between attributes of nodes.
To address these problems, we propose a parameter-free algorithm named CSPM (Compressing Star Pattern Miner) which identifies star-shaped patterns that indicate strong correlations among attributes via the concept of conditional entropy and the minimum description length principle.  Experiments performed on several benchmark datasets show that CSPM reveals insightful and interpretable patterns and is efficient in runtime. Moreover, quantitative evaluations on two  real-world applications show that CSPM has broad applications as it successfully boosts the accuracy of graph attribute completion models by up to 30.68\% and uncovers important patterns in telecommunication alarm data.  
\end{abstract}

\IEEEpeerreviewmaketitle

\section{Introduction}
\label{sec:introduction}
Data describing relationships between objects as a graph are abundant in many fields such as bioinformatics, chemistry, social network, and telecommunication~\cite{surveyDYN,surveyFSM,yang2020featurenorm,yang2021discrete}. 
To help on the understanding of graph data and support decision-making, various algorithms have been developed for mining interesting patterns~\cite{surveyDYN,surveyFIM,gspan}. Nonetheless, these algorithms have two main limitations:

First, most algorithms focus on extracting patterns from the topology (e.g., frequent subgraphs)~\cite{surveyDYN,surveyFSM}. However, they overlook the information carried by the attributes which is also important to understand the graph's properties. %
For example, in a social network, the multiple attribute values associated with users are more informative than the topological structure. 
Thus, uncovering the correlation among attribute values can further help to understand users' characteristics and assist for several tasks such as  user profile inference~\cite{dougnon2016inferring,chen2020learning}. A second example is  telecommunication networks where devices (vertices) are connected by telecommunication links (edges), and a device can raise various alarms or errors (attribute values). Discovering relationships between these alarms is  useful for network fault management to reduce the downtime and the maintenance cost~\cite{fournier2020discovering}.
 
 Second,  graph pattern mining algorithms typically require the setting of multiple parameters to obtain patterns~\cite{surveyFSM,fournier2019tkg,gSpan2002}. However, finding suitable values for those parameters is time-consuming and often unintuitive. For instance, in frequent subgraph mining, a parameter called minimum support must be set. The setting of this parameter is hard as it depends on the dataset's characteristics that are initially unknown to the user. 
 Finding suitable values for those parameters is typically difficult and usually done by trial and error. In fact, if parameters are not set properly, too few or too many patterns may be found, which may lead to miss important information or to find many spurious patterns.

To address the above limitations, this paper proposes an algorithm, named CSPM (Compressing Star Pattern Miner), for identifying representative patterns in attributed graphs.
\color{black}

These patterns, named \emph{attribute-stars} (\emph{a-star}), are star-shaped and designed to reveal strong relationships between attribute values of connected nodes, in terms of conditional entropy. 

An a-star indicates that if some attribute values  occur in a node, some other attribute values will appear in some neighbors of that node. 
This type of attribute patterns is simple yet meaningful for many domains. For instance, in social networks, an a-star gives the information on characteristics of a user and his/her friends (e.g., friends of a smoker also tend to be smokers).

Such patterns can be used to perform \emph{node attribute completion} (filling missing attribute values) \cite{chen2020learning}, completing user profiles based on their friends' information.

Note that a star shape is  desirable as more complex structures are generally less meaningful. For instance, it is known that the influence of the friends of a friend 
tends to be much weaker than that of 
direct friends~\cite{dougnon2016inferring}. Also,  patterns with complex links are hard to interpret.
\color{red}

\color{black}

CSPM is easy to use as it is a parameter-free algorithm. 
Besides, CSPM applies a greedy search to quickly find an approximation of the best set of patterns that maximize the compression according to the MDL (Minimum Description Length) principle. Moreover, CSPM also relies on the concept of conditional entropy to assess how strong relationships between attributes are. Experiments have been performed on several benchmark datasets and show that the proposed algorithm is efficient and reveals insightful and interpretable patterns. Moreover, it has been quantitatively verified that CSPM successfully boosts the accuracy of graph attribute completion models (e.g., GCN~\cite{kipf2016semi}, GAT~\cite{velivckovic2017graph}) and it can uncover important patterns in telecommunication alarm data.

The rest of this paper is organized as follows. Related work is reviewed in Section~\ref{sec:related work}.
Preliminaries are introduced in Section~\ref{sec:preliminaries}. CSPM  is described in Section~\ref{sec:CSPM}. Experiments are presented in Section~\ref{sec:experiments}. Finally, a conclusion is drawn in Section~\ref{sec:conclusion}.

\section{Related Work}
\label{sec:related work}

One of the most popular tasks to find patterns in a graph database or single graph is frequent subgraph mining (FSM)~\cite{surveyFSM}. 
It consists of finding all connected subgraphs having an occurrence count (support) that is no less than a user-defined minimum support threshold. 

However, finding large subgraphs is sometimes unnecessary for decision-making. 
Hence, special cases of the FSM problem that are easier to solve have been studied, such as mining frequent trees~\cite{attributetree,tree} and  paths~\cite{atzmueller2018minerlsd:,paths}. But many graph mining algorithms can only handle graphs with a single label per node. A few algorithms can find patterns in graphs with multiple labels per node (attributed graphs)~\cite{kargar2019mining}. Pasquier et al. proposed to mine frequent trees in a forest  of attributed trees~\cite{attributetree}, while Atzmueller et al. designed the MinerLSD algorithm to find core subgraphs in an attributed graph \cite{atzmueller2018minerlsd:}. 
Besides, algorithms were designed to find temporal patterns in dynamic attributed graphs~\cite{surveyDYN,aerminer}. 
However, most graph pattern mining approaches have many parameters that users generally set by trial and error, which is time-consuming and prone to errors. Unsuitable parameter values can lead to long runtime or finding too many patterns, including spurious ones.

\textbf{Mining compressing patterns.}
To select a small set of patterns that represent well a database, an emerging solution is to find \emph{compressing patterns}. This idea was introduced in the Krimp  algorithm~\cite{vreeken2011krimp:} for discovering interesting itemsets (sets of values) in a transaction database (a binary table).  Krimp  applies the MDL principle~\cite{mdl} based on the idea that the best model will be insightful as it represents key patterns of the data. A database is compressed by a model $M$ by encoding each occurrence of a pattern in the database by a code (stored in a code table). 
Though Krimp does not guarantee finding the best model, it typically finds a good one. However, the binary database format of Krimp is simple, which restricts its applications. To go beyond binary tables, variations of Krimp were proposed. SeqKrimp~\cite{lam2014mining}  was designed to mine compressing sequential patterns in a set of sequences and the DITTO~\cite{bertens2016keeping} algorithm was proposed to find compressing patterns in an event sequence.
  For graphs, GraphMDL was introduced to mine compressing subgraphs in a labeled graph~\cite{bariatti2020graphmdl}.  But GraphMDL requires that the user first discovers frequent subgraphs using a traditional FSM algorithm~\cite{surveyFSM} (similarly to Krimp). Hence,  GraphMDL is not parameter-free, and results may vary widely depending on how the user sets the FSM algorithm's parameters. 
  Besides, GraphMDL handles a single or few attribute(s) and is not easily generalizable to multiple attributes. Hence, it fails to model richer data such as the multiple attribute values of social network users.

Differently from these approaches, this study presents a parameter-free algorithm to mine compressing patterns in graphs. To avoid relying on a traditional pattern mining algorithm, the proposed algorithm adopts an iterative approach to find a good set of compressing patterns, which is inspired by an improved version of Krimp, named SLIM~\cite{smets2012slim:}. 
During each iteration, candidates are found on-the-fly rather than using a predetermined set of patterns. 
Another difference distinguishing the algorithm presented in this paper from the previous work is that it handles an attributed graph as input and it finds a pattern type called \emph{attribute-stars}. These patterns have a simple topological structure but they provide rich information about the relationships between attribute values of a node and its direct neighbors.

\textbf{Summarizing and compressing a graph.}
A related research area is techniques for summarizing and compressing graphs. They focus on reducing storage space for large graphs to ease their processing or understanding.  For example, SlashBurn~\cite{6807798} removes high degree nodes of a graph to create large connected subgraphs. Then, the adjacency matrix is reordered to achieve high compression.
Another method named VOG~\cite{koutra2014vog:}  decomposes a graph into subgraphs, and each subgraph is approximated using six pattern types (full/near clique, chain, full/near bipartite core, and star). The approximation having the MDL is selected.
Another algorithm called HOSD~\cite{ahmed2020interpretable} compresses a graph by searching for subgraph types of fixed size at multiple scales (from a local  to a global perspective). But like VOG, HOSD focuses on compressing a graph's topological structure and ignores node attributes.

To summarize a graph based on its topological and attribute structure,  Greedy-Merge~\cite{liu2018graph} adds virtual edges between nodes with same attribute values and then finds clusters of vertices that have a similar topological structure and attributes.
The QB-ULSH~\cite{khan2014set-based} algorithm compresses an attributed graph by replacing nodes with similar edges  by super-nodes using the MDL principle. 
However, the user needs to set multiple parameters, which directly influences results. 
Another recent approach is  ANets~\cite{amiri2018efficiently}. It summarizes a directed weighted attributed graph by replacing nodes and edges by super-nodes and super-edges, through a process of matrix perturbations. 
The CSPM algorithm proposed in this paper is different. It is parameter-free and  finds patterns that explain relationships between many attribute values of connected nodes. Though CSPM is compression-based, the compression is a mean to obtain representative patterns rather than the goal. 

\begin{remark}
\color{black}
The input data, output patterns, and the focus of CSPM are quite different from GraphMDL and VOG (see Table~\ref{tab:comparison}). VOG focuses on summarizing the topology of patterns without attributes. GraphMDL is designed to find patterns in many small graphs instead of a large one. 
SLIM has a goal and procedure similar to CSPM since candidates are found on-the-fly rather than using a pre-determined set of candidates, but it can't handle attributed graphs. 
\color{black}
\end{remark}

\begin{table}
\footnotesize
\centering
\caption{{\color{black}Comparison between CSPM and  related work}}
\label{tab:comparison}
 \resizebox{0.49\textwidth}{!}{
\begin{tabular}{lccccc} 
\toprule
                            & CSPM & Krimp & SLIM & GraphMDL & VOG  \\ 
\midrule
Attributed graph?    &  \Checkmark & \XSolidBrush & \XSolidBrush  & \XSolidBrush & \XSolidBrush    \\
Atribute patterns?  &  \Checkmark  & \XSolidBrush  & \XSolidBrush & \XSolidBrush        & \XSolidBrush    \\
Compressing patterns? & \Checkmark & \Checkmark & \Checkmark  & \Checkmark        & \XSolidBrush    \\
On-the-fly candidates?     & \Checkmark & \XSolidBrush  & \Checkmark  & \XSolidBrush        & \XSolidBrush    \\
\bottomrule
\end{tabular}}
\end{table}

\section{Preliminaries}
\label{sec:preliminaries}

In this section, important definitions related to graphs and the discovery of compressing patterns are presented. 

\begin{figure}[h]
\vspace{-5pt}
\centering
\includegraphics[width=0.97\columnwidth]{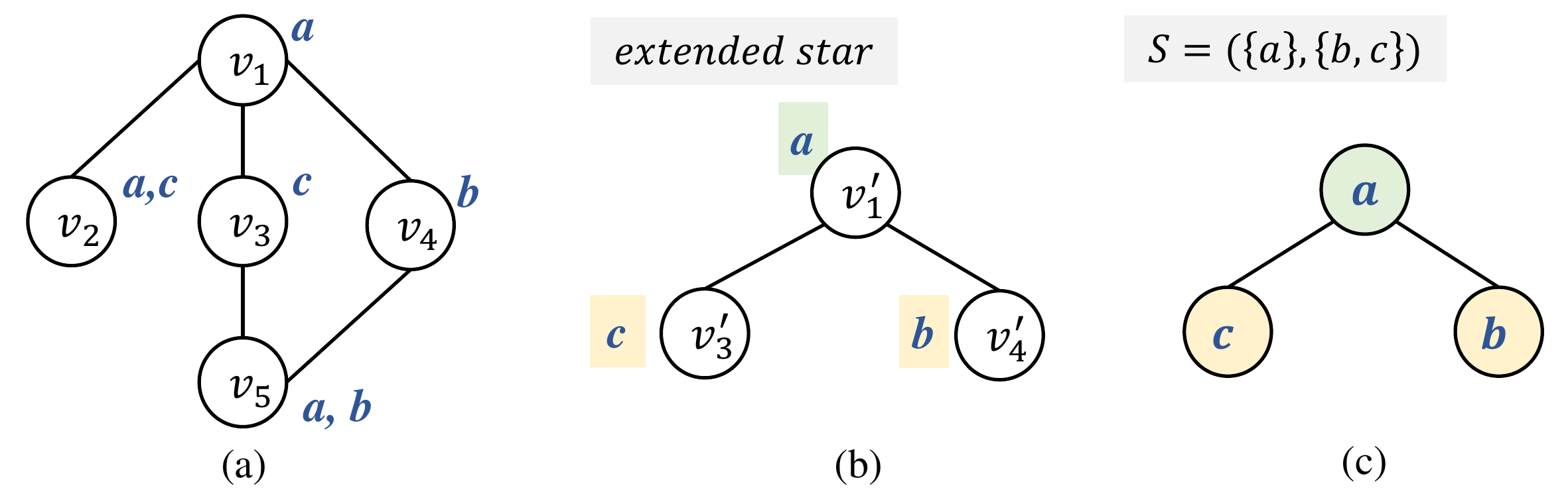}
\caption{An example of attributed graph}
\label{fig:toy_graph}
\vspace{-8pt}
\end{figure}

\textbf{Graphs.} A \emph{graph} $G = (V, E)$ is composed of two finite sets: a vertex set $V$ and an edge set $E$. The vertex set $V$ is a set of one or more elements called \emph{vertices}. The edge set has zero or more elements called \emph{edges}, where $E \subseteq V \times V$.
A vertex $u$   \emph{can reach} and is \emph{adjacent to} a vertex $v$ if $\{u,v\} \in E$. An edge $e_1$ is adjacent to an edge $e_2$ if they contain the same vertex.
A graph $G$ is \emph{connected} if any vertex $u$ can be reached from any other vertex $v$ by traversing a sequence of adjacent edges (a \emph{path}). 
An \emph{attributed graph}  
$G = (A, \lambda, V, E)$ is 
a set $V$ of vertices, 
a set  $E$ of edges,
a set of nominal attributes $A$,
and a relation  $\mathcal\lambda: V \mapsto A$ that maps vertices to attribute values. Attributed graphs can model data from many domains such as social networks where nodes are persons, edges are relationships between persons, and attribute values describe a person's profile in terms of properties such as age, gender, and city. In this paper, the input is an attributed graph with nominal attributes, that is connected and does not contain self-loops (and edge from a vertex to itself).  For example, Fig.~\ref{fig:toy_graph}(a) shows an attributed graph that will be the running example. It has five vertices $V= \{v_1, v_2 ,\ldots, v_5\}$, three attribute values ($a$, $b$ and $c$). For instance,  $v_2$ has attribute values $a$ and $c$.

A graph can be represented as a \emph{vertex adjacency list}, i.e. 
a list of tuples, where each tuple contains a vertex followed by the set of its adjacent vertices. For example, the adjacency list of the graph of Fig.~\ref{fig:toy_graph} (a) is 
$\{
(v_1, \{v_2, v_3, v_4\}),$ $
(v_2, \{v_1\}),$ $
(v_3, \{v_1, v_5\}),$ $
(v_4, \{v_1, v_5\}),$ $
(v_5, \{v_3, v_4\})
\}$.

Many types of structural patterns were found in graphs in previous work such as trees, paths, cliques, bi-partite cores, stars, and  other complex subgraphs~\cite{surveyDYN,surveyFIM,koutra2014vog:}.
 A \emph{star} is a  graph where a vertex $w$ called the \emph{core} is adjacent to all other vertices (called \emph{leaves}) and no leaves are adjacent to each other.
Formally, a \emph{star} $X$ is an undirected graph $X = (V,E,L,c)$ where
$V$ is a set of vertices,
$c \in V$ is a vertex called the \emph{core},
$L = V - \{c\}$ is a non empty set of vertices called \emph{leaves},
and there are edges between the core and each leaf, i.e.
$E = \{ \{c,w\} | w \in V \wedge  w \not = c \}$.

{\color{black}
The definition of star does not consider attributes. To add attributes, the concept of star is extended as follows: An \emph{extended star} $X = (V,E,L,c, A, \lambda)$ is a star with a set of attribute values $A$ and a function  
$\mathcal\lambda: V \mapsto A$ that maps vertices to categorical attribute values.
To be able to locate occurrences of an extended star in an attributed graph,  the concept of appearance is defined. An extended star $X = (V_x,E_x,L,c,A_x, \lambda_x)$ is said to \emph{appear in} an attributed graph $G = (A_y, \lambda,_y V_y, E_y)$ if there exists a bijective mapping $f:V_x \rightarrow V_z$ (where $V_z \subseteq V_y$) that meets two conditions:
First, for each pair $(v,a) \in \lambda_{x}$, there exists a corresponding pair  $(f(v),a) \in \lambda_{y}$.
Second, for each edge $(u,v) \in E_x$,
there is a corresponding edge $(f(u),f(v)) \in E_y$.
}

\textbf{Compressing Patterns.} The basic framework for mining compressing patterns was defined in  Krimp~\cite{vreeken2011krimp:}.
The aim is to discover the model $M$ (a set of patterns) that best describes the database. The quality of a model is evaluated using the Minimum Description Length (MDL) principle on the basis that a model that is small and compresses well the database will capture the key information that the database contains.

In Krimp, a model $M$ is composed of a set of patterns (sets of items, i.e. itemsets) and their corresponding code lengths. To compress a database, each itemset $X$ of a model is replaced by a unique code. Patterns that appear frequently are given smaller codes to maximize compression. In Krimp, the length of a code for an itemset $X$ is represented as a number of bits and is calculated using the Shannon Entropy as $L(X) = - log_2{P(X)}$, where $P(X)$ is the relative occurrence frequency of $X$. Before building a model, Krimp creates a two-column table called \emph{Standard Code Table} ($ST$) where each line stores an item from the dataset in the left column and its corresponding code length in the right column. Then, the length of the original database $L(D|ST)$ without compression can be calculated by replacing all items in the database by their codes in $ST$ and summing their code lengths with that of the $ST$.
To build a better model $M$, Krimp initializes $M = \emptyset$ and then iteratively adds a pattern to the model if it improves compression.
Each pattern is an itemset consisting of one or more items.
A model is represented as a code table $CT$ assigning a code length to each pattern.
The description length (DL) of the database compressed with $CT$ is calculated as $L(CT,D) = L(CT|D) + L(D|CT)$, where $L(CT|D)$ is the length of the code table, and $L(D|CT)$ is the length of the database encoded with that code table. To find a good model (a code table), Krimp employs a greedy search. 
{\color{black} Note that the ultimate goal of compressing pattern mining algorithms is to find a good set of patterns that describe the data well,
instead of actually compressing the data.
Hence, only the code length of each pattern is necessary, which can be directly obtained by well derived formulas (e.g. Shannon Entropy).}

\section{The CSPM Algorithm}
\label{sec:CSPM}

This section describes the proposed CSPM algorithm to discover star-shaped attribute patterns revealing interesting relationships between attributes in an attributed graph. 

\subsection{Pattern Format {\color{black}and Problem Statement}}

The goal of this study is to mine \textbf{attribute} patterns that can reveal strong inner relationships between attribute values with implicit connection information, instead of finding subgraphs or substructures meeting the given requirements(e.g. frequency).  
With this premise, a star-shaped pattern format is selected. 
This format has some advantages. First of all, a star-shaped attribute pattern is simple, easy to understand, and can indicate correlation between attribute values of  directly connected nodes (illustrated by edges). 
What's more, for practical applications such as social network analysis, a star-shaped attribute pattern can capture the influence relationship between a person and its friends~\cite{shih2020data}. This pattern format also has industrial application for network alarm analysis as it will be shown in Section \ref{sec:experiments}.

Although extended stars {\color{black} (defined in Section~\ref{sec:preliminaries})} could reveal interesting patterns, a more general type of patterns is defined to summarize multiple extended stars. The goal is to focus less on the structure of stars and more on the relationships between attribute values of cores and leaves. The proposed pattern type $S = (S_c, S_L)$, namely \emph{attribute-star} (\emph{a-star}), where $S_c$ is a set of attribute values of a core node
and $S_L$ is a set of attribute values that appear in any of its leaf nodes.
The sets $S_c$ and $S_L$ of an attribute-star are  called \emph{coreset} and \emph{leafset}, and their elements, \emph{core values} and \emph{leaf values}, respectively.

The relationship between an attribute-star and a star is defined as follows: An attribute-star $S = (S_c, S_L)$ is \emph{matching with} a star $X = (V,E,L, c, A, \lambda)$, if  (1) $\forall a \in S_c$, there is a pair $(c,a) \in \lambda$ and (2)  $\forall y \in S_L$, there is a leaf $u \in L$ such that $(u, y) \in \lambda$. Thus, an attribute-star can match with multiple stars which gives the possibility to summarize them.
For example, consider the a-star $S=(\{a\},\{b,c\})$ in Fig.~\ref{fig:toy_graph}(c). It indicates that a core has the attribute value $a$ and some leaves have attribute values $b$ and $c$. This a-star matches with a star depicted in Fig.~\ref{fig:toy_graph}(b).

In the proposed method, the role of core values within an a-star is different from that of leaf values. More precisely, the core values are viewed as influencing  leaf values. In other words, for each a-star that is discovered, if core values appear in the core vertex, there is a high chance that the leaf values appear in its neighbor(s).

 {\color{black}\textbf{Problem statement.} This study aims to mine compressing patterns in an attributed graph. The \textit{input} is an attributed graph $ G=(A, \lambda, V, E)$ with categorical attribute values. The \textit{goal} is to find the set of a-stars $\{S_1, S_2, \cdots, S_n\}$ 
 that best compresses the original information of the attributed graph $G$ losslessly, i.e, with the minimum description length. The \textit{output} is a set of a-stars ordered by ascending code lengths. An a-star with a shorter code length indicates that it is more informative.
 
 Note that we aim to find a subset of a-stars that meets the above  \emph{goal} instead of finding \textit{all} possible a-stars. For this problem, an approximate algorithm is desired since the brute force approach is impracticable. For instance, if only 500 a-stars exist, there is $2^{500}-1$ possible subsets.} 
 
\subsection{The Inverted Database Representation}
To support the efficient discovery of a-stars, the CSPM algorithm transforms the input database into an \emph{inverted database} ($I$) representation. This latter allows to easily find a-stars by merging lines representing smaller a-star patterns together. Finding a-stars that best compress the database using the MDL then becomes a problem of selecting appropriate pairs of lines to be merged to make new a-stars.

The inverted database representation is based on the observation that an attributed graph  contains two parts: The topology (a vertex adjacency list) and the mapping function between vertices and attributes.
Each tuple in the adjacency list can be viewed as a star associating a vertex (the core) to a list of adjacent vertices (the leaves). For example, the tuple $(v_1, \{v_2, v_3, v_4\})$ in the adjacency list of the graph of Fig.~\ref{fig:toy_graph}(a) is a star.
Note that any vertex from a graph can be the core of a star. 
For that reason, a-stars can be directly found by substituting the vertices in the adjacency list with their corresponding attribute values.
For instance, the tuple $(v_1, \{v_2, v_3, v_4\})$ could be used to form the a-star $(\{a\}, \{a,b,c\})$.
Considering the fact that the core in an a-star has a different role than the leaves, each attribute value can be labeled as a core value or leaf value according to the vertices where it appears.

The inverted database is a three-column table. The first column  $S_L$ contains   leafsets, the second column $S_c$ indicates the coresets that are connected to the corresponding leafsets in the first column to form an a-star, and the third column contains the set of vertices where the core values appears, which are called the \emph{positions}. 
A mapping table can be also built to map each core to its list of positions.

\begin{figure}[h]
\vspace{-10pt}
\centering
\includegraphics[width=0.44\textwidth]{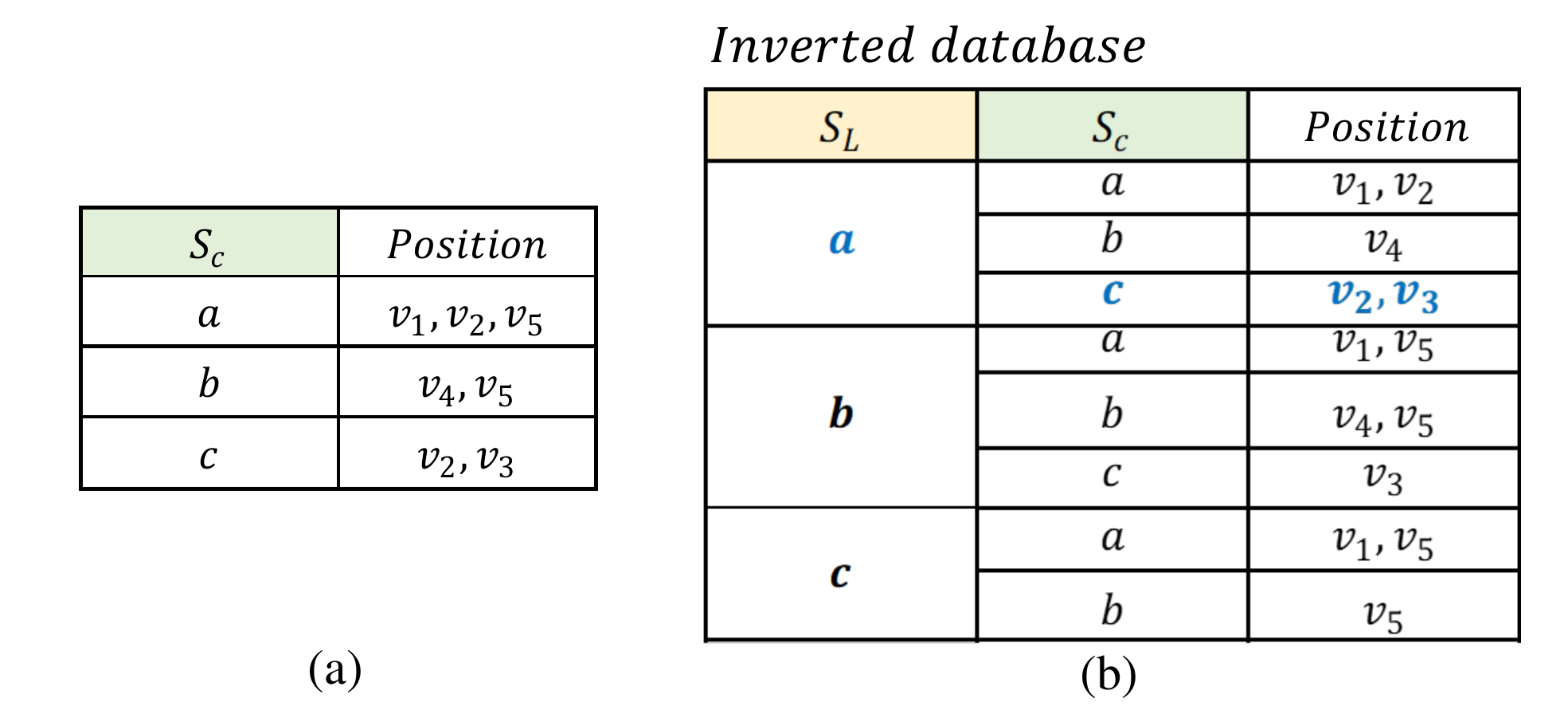}
\caption{Mapping table and inverted database}
\label{fig:invertedDatabase}
\vspace{-1pt}
\end{figure}

Taking the graph depicted in Fig.~\ref{fig:toy_graph} as example, the mapping table of this graph is shown in Fig.~\ref{fig:invertedDatabase}(a), and the inverted database generated from this graph is presented in  Fig.~\ref{fig:invertedDatabase}(b).  From this mapping table, it can be seen that the coreset $S_c:\{c\}$ appears at vertices $v_2$ and $v_3$. Moreover, from the graph, it can be observed that $\{a\}$ is adjacent to $\{c\}$ in both cases at $v_2$ and $v_3$. Accordingly, there is a blue record \{$\{a\}$,$\{c\}$,$\{v_2,v_3\}$\} in the inverted database of Fig.~\ref{fig:invertedDatabase}(b). 

There are several advantages of using this inverted database representation over covering attribute patterns directly on the original graph:

First, it is easy to calculate the code length of each a-star. The reason is that the co-occurrence frequency of leaf values that are adjacent to the same core value can be directly obtained by intersecting their position sets.
Second, there is no need to scan the whole database again after the discovery of each new a-star pattern. 
This is because each line of the database is initialized as a basic a-star with one edge (core value, leaf value). Hence, all a-stars can be generated by merging two or more lines together.
Third, the inverted database allows storing a-stars in a non-redundant way where each line of the inverted database represents a distinct a-star. 
Thus, it is convenient to encode each a-star by appending a distinct code for each line, instead of building an extra translation table.

\subsection{MDL with the Inverted Database.}

The CSPM algorithm is inspired by the basic framework for mining compressing patterns of  Krimp~\cite{vreeken2011krimp:}. 
Krimp relies on a two-column standard code table $ST$ and a code table $CT$ to calculate the total description length of the compressed database based on the MDL principle. 
But the data and pattern types considered by Krimp are simple as it only finds itemsets (sets of symbols) in a transaction database (a binary table)~\cite{vreeken2011krimp:}. 
In this paper, the goal is to discover a-stars in an attributed graph, which is more complex as patterns have two parts which are core values and leaf values.

Therefore, the Krimp model is adapted to define a model $M$ that is suitable for a-star patterns.
The proposed model type contains two code tables, named $CT_c$ and $CT_L$. The former is a traditional two-column code table used to encode coresets, while $CT_L$ is used to encode leafsets. This latter contains pointers to $CT_c$ to facilitate appending the codes of coresets and leafsets to calculate the description length. 
$CT_L$ has a similar structure to  the inverted database. The difference is that the third column of $CT_L$ contains  leaf values' codes instead of sets of positions.   

Before explaining how CSPM  calculates the DL, it is important to notice that it is necessary  to first build the \emph{standard code table} $ST$ based on the frequencies of all attributes in the attributed graph.
The standard code table is the optimal encoding of all attributes without labels and structure information. It contributes to the cost (description length) of coresets and leafsets  stored in code tables. 
In the case where patterns have a single core value, the format of the code table $CT_c$ is the same as that of the standard code table.  

\begin{figure}[h]
\vspace{-5pt}
\centering
\includegraphics[width=0.48\textwidth]{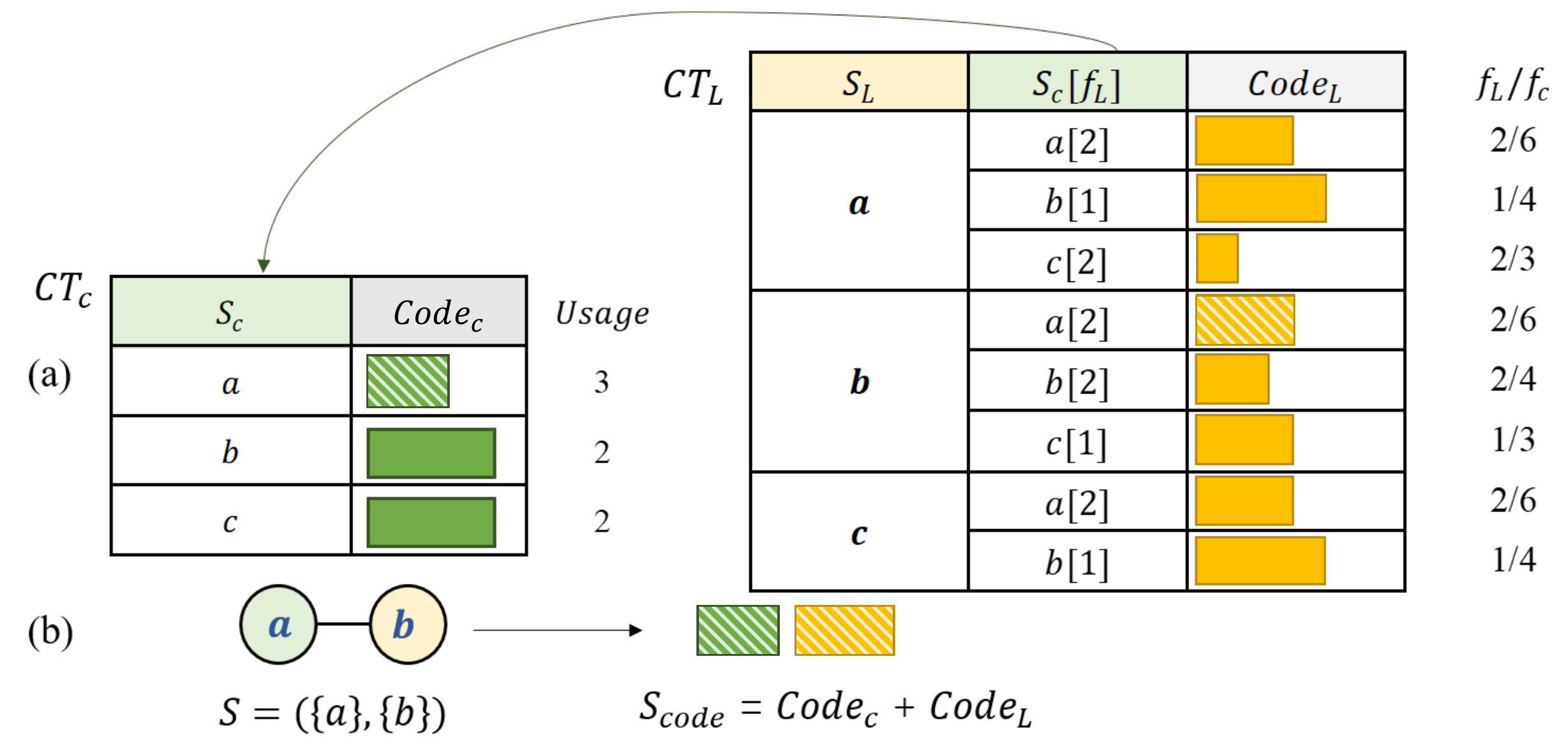}
\caption{Two code tables and an a-star $S=(\{a\},\{b\})$ encoding}
\label{fig:code table}
\vspace{-5pt}
\end{figure}

For the sake of simplicity, we first discuss how to mine a-stars having a single attribute value as core values. Then, the more general case of multiple core values will be explained in subsection~\ref{subsec:detailed}.
In the case of single-core value patterns, a-stars are found by simply finding the pattern set of $S_L$ adjacent values for  different core values to form the different a-stars.
As example, Fig.~\ref{fig:code table}(a) shows the $CT_c$ and $CT_L$ code tables for the graph depicted in  Fig.~\ref{fig:toy_graph}.
In that figure, numbers under the label \emph{Usage} indicate the frequency of $CT_c$ in the attributed graph derived from the mapping function table. Note that this usage information is not a part of the $CT_c$ table. 
In $CT_L$, the first two columns can be obtained by combining the corresponding codes of $Code_c$ in $CT_c$. Moreover, the column $Code_L$ records all codes of leafsets $S_L$ adjacent to different coresets. 
To obtain the code of an a-star, a pointer from the column $S_c$ to the corresponding core values $S_c$ in $CT_c$ is used.

In Fig.~\ref{fig:code table}(a), values are provided in a column under the label $f_L/f_c$. The $f_L$ means the frequency of the corresponding a-star (line) in $CT_L$, while $f_c$ is that of the corresponding $S_c$ in the inverted database. Note that, the column $f_L/f_c$ is not a part of $CT_L$.
Consider the a-star $(\{a\}, \{b\})$ as an example. This a-star can be represented by Fig.~\ref{fig:code table}(b). 

The description length $L(M,I)$ of a model and inverted database is then calculated by the following equations as Equation~\eqref{eq: MDL}:

\vspace{-3pt}
\begin{equation}
\small
    L(M,I) = L(M) + L(I|M).
\label{eq: MDL}
\end{equation}
\begin{equation}
\small
    L(M) = L(CT_c|I) + L(CT_L|I).
\label{eq: model length}
\end{equation}

Note that, the calculation of $L(M)$ seen as Equation~\eqref{eq: model length} is similar to the equation used by the Krimp algorithm.  It is calculated by adding the cost of all columns together. 

The length $L(I|M)$ is derived directly from the proposed inverted database representation by Equation~\eqref{eq:code length of I}: 

\begin{equation}
\small
    L(I|M) = \sum_{S \in I_{new}} L(S_{code})*S(f_L),
\label{eq:code length of I}
\end{equation}

\begin{equation}
\small
    L(S_{code}) = L(Code_c) + L(Code_L).
\label{eq:code length}
\end{equation}

where $S(f_L)$ is the frequency of the a-star (line) $S$ in the new inverted database $I_{new}$, which is described using $M$. And $L(S_{code})$ is the code length of $S$ by summing up the code lengths of its coreset $Code_c$ and leafset $Code_L$.

The CSPM algorithm is designed to find a-stars that indicate strong relationships between core values and leaf values according to the Minimum Description Length (MDL) principle. That is, CSPM is designed to find a-stars meet the following two conditions. First, an interesting a-star is supposed to be frequent to some extent. And the second condition is that the leaf values need to have high relevance for some core values while being less relevant to other core values. Relevance is assessed through the conditional entropy, described next.

\subsection{Conditional Entropy}

The code length of an a-star is required to calculate the DL of a model. The proposed method differs from traditional methods such as Krimp that are based on the Shannon entropy to obtain optimal lengths. There are two reasons. First, the proposed model considers two entities (based on core and leaf values), while Shannon entropy is only suitable for the case of one variable. Second, our goal is to evaluate the relevance between leaf values and core values instead of merely focusing on the frequency.

Calculating the code $Code_c$ of core values in $M$ is only based on the frequency of attribute values in the mapping function between vertices and attribute values, as Equation~\eqref{eq:Enropy}. To optimally encode leaf values connected to each coreset

in each a-star, conditional entropy is used as a key measure.

\vspace{-8pt}
\begin{equation}
\small
    L(cv) = -\log (P(cv)).
\label{eq:Enropy}
\vspace{-5pt}
\end{equation}
\vspace{-10pt}

In information theory, the conditional entropy $H(Y|X)$ measures the cost of describing a random variable $Y$ supposing that the value of another random variable $X$ is given \cite{cover1999elements}. 
Similarly, in this paper, the set of coresets $S_c^M$ can be regarded as $X$ and the set of leafsets $S_L^M$ as $Y$ which can be calculated given that the coreset $S_c$ have been obtained. 
Based on this fact, the relevance of each coreset to its leafset (a set of leaf values) can be revealed using the conditional entropy.
According to the principle of  conditional entropy, the code length of each a-star (each line in $CT_L$) with leafset $S_L$ and coreset $S_c$ is given by Equation~\eqref{eq:length of each line}. 
Hence, the $Code_L$ for each line is only determined by the corresponding $f_L$ and $f_c$ values.

\vspace{-2pt}
\begin{equation}
\small
\begin{split}
       L(Y = S_L| X = S_c) = -\log {\frac{p(S_L, S_c)}{p(S_c)}}  
       = -\log \frac{f_L}{f_c}.
\end{split}
\label{eq:length of each line}
\end{equation}
\vspace{2pt}

Then, $H(Y | X)$ is calculated by Equation~\eqref{eq:conditonal_entropy}, which  represents the average encoding cost of each line in $CT_L$ after using the conditional entropy encoding method.

\begin{equation}
\small
\begin{split}
    H(Y|X) = -\sum_{S_c\in X,S_L \in Y} {p(S_c,S_L)\log \frac{p(S_c,S_L)}{p(S_c)}}  \\
   = -\sum_{j=1}^m {\sum_{i=1}^n {\frac{l_{ij}}{s} \log \frac{l_{ij}}{c_j}}}\\
\end{split}
\label{eq:conditonal_entropy}
\end{equation}

In that equation, $s$ is the total frequency of a-stars (lines) in the inverted database, which is the sum of $f_L$, $m$ is the total number of coresets, $n$ is the total number of a-stars (lines) in the inverted database, $l_{ij}$ is the frequency $f_L$ of the $i_{th}$ leafset and $j_{th}$ coreset, and $c_j$ is the frequency $f_c$ of the $j_{th}$ coreset.

Based on the above equation, the total encoding cost of the inverted database is obtained by Equation~\eqref{eq: total encode length}. Note that $\sum_{i=1}^n l_{ij}$ is equal to the frequency of $c_j$. 

\begin{equation}
\small
\begin{split}
    L(I|M) = -s \times H(Y|X) = \sum_{j=1}^m {\sum_{i=1}^n l_{ij} \log \frac{l_{ij}}{c_j}} \\
    = \sum_{j=1}^{m} \sum_{i=1}^n l_{ij} \log{c_j} - \sum_{j=1}^m \sum_{i=1}^n l_{ij}\log{l_{ij}} \\
    = \sum_{j=1}^{m} c_j\log{c_j} - \sum_{j=1}^m \sum_{i=1}^n l_{ij}\log{l_{ij}}.
\end{split}
\label{eq: total encode length}
\end{equation}

\subsection{Candidate Generation}
The model is built iteratively. After determining all the coresets, the problem is simplified into selecting the leafsets with minimal conditional entropy.
An intuitive way of doing this is to  select the best set of a-stars among all possible combinations of leafsets with each coreset. But this is time-consuming for large databases.
To generate patterns without assuming that a set of candidates has been previously mined by another algorithm as in Krimp, CSPM uses a greedy approach that merges the two leafsets that provide the best gain $\Delta L$, i.e, in terms of the difference between the DLs before/after merging.
The gain of a pair represents how much the merge step will reduce the DL. Thus, a larger gain is better.

Suppose that $p$ 
is the pair of leafsets $\{(S_{Lx},S_{Ly})|$ $S_{Lx} \in CT_L,$ $ S_{Ly} \in CT_L\}$ that CSPM is going to merge.
A merge operation means that a new leafset pattern $\{S_{Lx} \cup S_{Ly}\}$ will cover all situations where $S_{Lx}$ and $S_{Ly}$ appear around the same coreset. This co-occurrence information is easily obtained using the inverted database by intersecting all sets of \emph{positions} of the two merged lines that have the same $S_c$.

Two parts influence the DL to generate a new pattern $\{S_{Lx} \cup S_{Ly}\}$. 
The first one is the cost increase of the new pattern's leafset in the code table, which can be easily obtained through the standard code table $ST$.
The other one is the gain $\Delta L$ from merging the pair of leafsets. From Equation~\eqref{eq: total encode length}, $L(I|M)$, Equation~\eqref{eq:gain} is obtained for a simplified calculation of  $\Delta L$: 

\begin{equation}
\small
\begin{split}
    \Delta L = \underbrace{(\sum_{i=1}^m c_j\log c_j - \sum_{i=1}^m c_j^{'}\log c_j ^ {'})}_{\text{$P_1$}} - \\
    \underbrace{(\sum_{j=1}^m \sum_{i=1}^n l_{ij}\log l_{ij} - \sum_{j=1}^m \sum_{i=1}^{n^{'}} l_{ij}^{'}\log l_{ij}^{'})}_{\text{$P_2$}}
\end{split}
\label{eq:gain}
\end{equation}

where variables with prime as superscript indicate the properties of the new code table $CT_L^{'}$ after the merge operation. Note that the number of coresets $m$ does not change during the process because the set of coresets $S_c^M$ is known.

To better explain the above equation, a detailed analysis of its two parts $P_1$ and $P_2$ is given. It can be observed that $P_1$ is related to coresets while $P_2$ is related to leafsets. In the following, for the sake of brevity, the lower case variable $x$ is used to denote the $f_L$ of the leafset $S_{Lx}$, i.e. $x = S_{Lx}.f_L$. Similarly, the variable $y$ is used to denote $S_{Ly}.f_L$. 

It can be observed that not all frequencies for each line $f_L$ and each coreset $f_c$ are changed by a merge operation. Suppose $C$ is the coresets that $S_{Lx}$ and $S_{Ly}$ are both connected with. For each coreset $e \in C$, $f_e$ is the frequency of coreset $e$, and $xy_e$ is the co-occurrence frequency of $S_{Lx}$ and $S_{Ly}$. By the convenience of the inverted database structure, $xy_e$ is obtained by intersecting the \emph{positions} of a-stars $(e, S_{Lx})$ and $(e, S_{Ly})$ directly. Especially, if $xy_e$ is equal to zero, the two lines can't be merged and the gain is equal to zero.
Thus, $P_1$ in Equation~\eqref{eq:gain} can be derived as in Equation~\eqref{eq:P1}:

\begin{equation}
\small
P_1 = \sum_{e \in C} (f_e\log f_{e} - (f_{e} - xy_{e})\log (f_{e} - xy_{e})).
\label{eq:P1}
\end{equation}

Due to the fact that not all values of $l_{ij}^{'}$ will become different from $l_{ij}$ after merging, the $P_2$ part is calculated by only summing up all relative merged lines using Equation~\eqref{eq:P2}, where, $P_{e}$ means the gain for generating a new a-star $(e, S_{Lx} \cup S_{Ly})$ by merging lines with core value $e$. 

\begin{equation}
\small
    P_2 = \sum_{e \in C} P_{e}. \\
\label{eq:P2}
\end{equation}

There are three cases to be considered according to the relationship between $xy_e$, $x_e$ and $y_e$. 

(1) \textbf{Partly merged}. In this case, the leafsets share some positions but each leafset has some unique positions,

i.e., $ xy_{e} \neq x_{e}, xy_{e} \neq y_{e}$. The formulation $P_e$ can be derived as Equation~\eqref{eq:Situation1}.

\begin{equation}
\small
\begin{split}
    P_e = (x_e\log x_e + y_e \log y_e)- [(x_e-xy_e) \log (x_e-xy_e) \\ +  (y_e-xy_e) \log (y_e-xy_e) + 
    xy_{e} \log xy_e ]. \\
\end{split}
\label{eq:Situation1}
\end{equation}
\vspace{1pt}

(2) \textbf{Two lines totally merged}. In this case, $xy_e = x_e$ and $xy_e = y_e$. The two lines (a-stars) are merged as a new pattern $(\{e\}, \{S_{Lx} \cup S_{Ly}\})$ because they always appear at the same positions. After merging, there is no a-star $(\{e\}, \{S_{Lx}\})$ or $(\{e\}, \{S_{Ly}\})$ in the inverted database or code table anymore. Thus, $P_e$ can be calculated as Equation~\eqref{eq:Situation2}. 

\begin{equation}
\small
\begin{split}
    P_{2_{e}} = (x_e \log x_e + y_e \log y_e) - (xy_e \log xy_e)
    =xy_{e} \log xy_{e}.
\end{split}
\label{eq:Situation2}
\end{equation}

(3) \textbf{One line totally merged}. Only one a-star will be removed by merging. In this case, $xy_e = x_e, xy_e \neq y_e$ or $xy_e \neq x_e, xy_e = y_e$, as shown as Equation~\eqref{eq:Situation3} and~\eqref{eq:Situation4}, respectively.

\begin{equation}
\small
\begin{split}
    P_e = (x_e \log x_e + y_e \log y_e) - (y_e \log y_e + xy_e \log xy_e)\\
    =y_e \log \frac{y_{e}}{y_{e}-xy_{e}} + xy_{e} \log (y_{e} - xy_{e}).
\end{split}
\label{eq:Situation3}
\end{equation}

\begin{equation}
\small
\begin{split}
        P_e = (x_e \log x_e + y_e \log y_e) - (y_e \log y_e + xy_e \log xy_e)\\
        x_{e} \log \frac{x_{e}}{x_{e}-xy_{e}} + xy_{e} \log (x_{e} - xy_{e}).
\end{split}
\label{eq:Situation4}
\end{equation}
\vspace{2pt}

\begin{figure}[h]
\small
\centering
\includegraphics[width=0.40\textwidth]{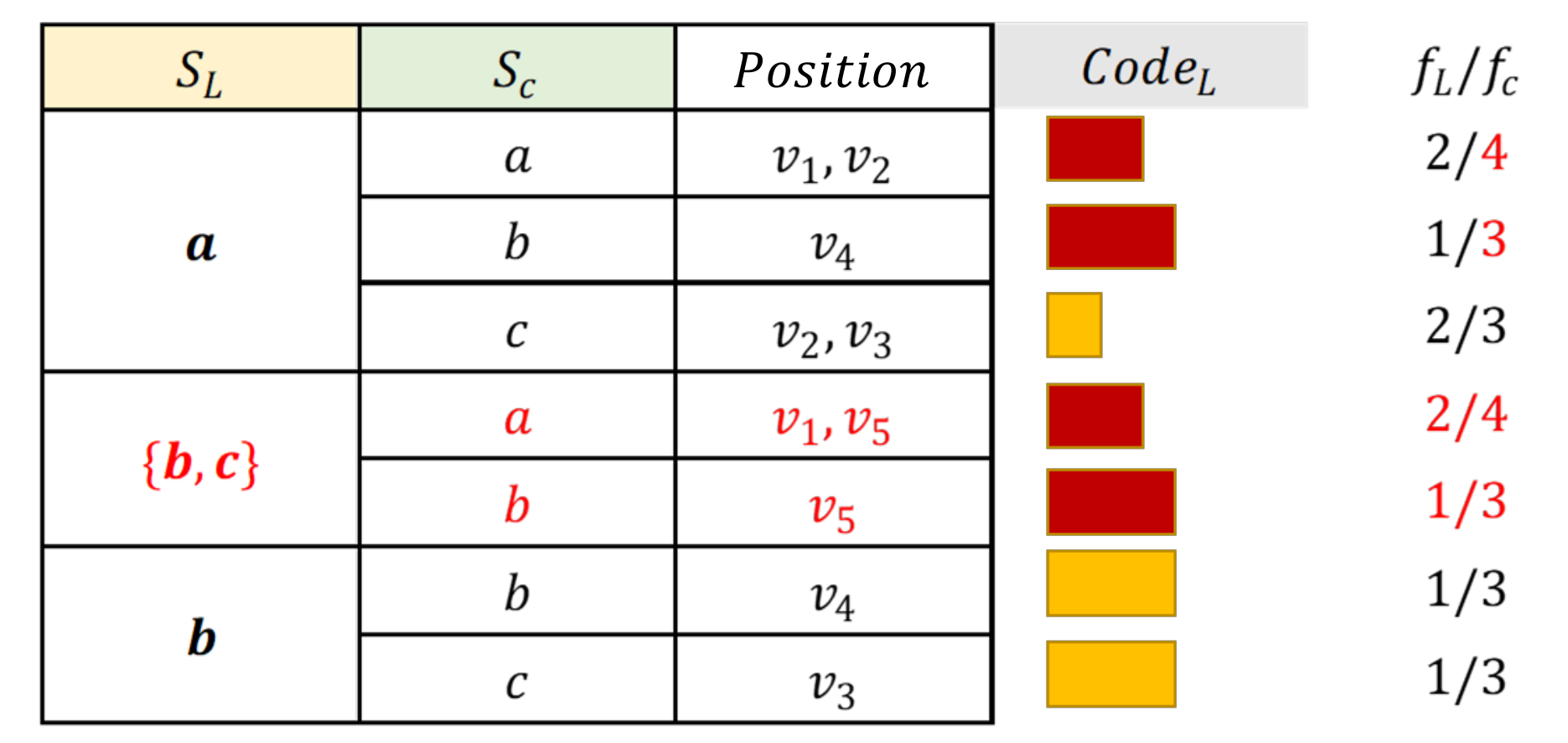}
\caption{Inverted database and $Code_L$ after merging $S_L$ pairs $(\{b\},\{c\})$}
\label{fig:afterMerge}
\vspace{-4pt}
\end{figure}

Take Fig.~\ref{fig:invertedDatabase} as example. Suppose that the leafsets $\{b\}$ and $\{c\}$ of column $S_L$ are merged. There are two coresets $C=\{\{a\},\{b\}\}$ that these leafsets are both connected with. For coreset $\{a\}$, the $Position$ of a-stars $(\{a\},\{b\})$ and $(\{a\},\{c\})$ are the same as $\{v_1,v_5\}$, which means the two lines can be totally merged as a new pattern (Case 2). And for $\{b\} \in C$, the common position is $\{v_5\}$. Hence, a-star $(\{b\},\{c\})$ will be totally merged  while the line  $(\{v_4, v_5\}-\{v_5\}=\{v_4\})$ remains for a-star $(\{b\},\{b\})$ (Case 3).
The new inverted database and code length $Code_L$ after merging are shown in Fig.~\ref{fig:afterMerge}. 
It is observed in Fig.~\ref{fig:afterMerge} that leafsets $\{b\}$ and $\{c\}$ are both connected to the same coreset $\{a\}$ of vertices $\{v_1,v_5\}$ and that they are connected to coreset $\{b\}$ of vertex $\{v_5\}$. 
Interestingly, the code length of leafset $\{b,c\}$ in $CT_L$ corresponding to core value $\{a\}$ is much shorter than those of previous $(\{a\},\{b\})$ and $(\{a\},\{c\})$. Overall, the database is compressed and its DL is reduced by the merge operation.

\subsection{The Detailed Algorithm}
\label{subsec:detailed}
The proposed CSPM algorithm takes as input an attributed graph $G$ (topology and mapping function) and outputs a set of compressing a-star patterns.  Algorithm~\ref{alg:basic} shows the detailed procedure.
It is divided into two sub-procedures for (1) determining and encoding all coresets $S_c^M$, (2) and constructing the final a-stars by finding  an approximation of the best leafsets for each coreset using a greedy search. The main steps of CSPM are:

\textbf{Step 1: Encode attribute values and core values}.

CSPM reads the mapping function without topology information, where each attribute value appearing at each vertex is seen as an item (symbol).  
If the user wishes to  mine  a-stars with coresets containing more than one core values, a traditional  compressing pattern mining algorithm can be applied on a transaction database composed of the attribute values of vertices. Several algorithms can be used such as Krimp \cite{vreeken2011krimp:} and SLIM \cite{smets2012slim:}. In the other case, attribute values are optimally encoded according to their global frequencies.
The result is a standard code table $ST$ of attribute values and a code table $CT_c$ for coresets $S_c^M$. Note that $CT_c$ is exactly the standard code table $ST$ if all coresets have one core value.

\textbf{Step 2: Construct the inverted database.} The inverted database $I$ is created, which contains leaf values $S_L$, core values $S_c$, and the topology information between core values and leaf values. Besides, each line in the initial inverted database stores the positions of each a-star with only one leaf value. The inverted database makes it easy to generate merge candidates and to cover the database with new patterns.

\textbf{Step 3: Generate leafsets merge candidates.} Merging  candidates allows showing the quality of leafset pairs that can be merged as a new pattern. To do so, it is necessary to calculate the gains $\Delta L$ of all possible pairs of leafsets in the inverted database. Then, pairs with $\Delta L > 0$ are sorted in  descending order to form the final candidates. 

\textbf{Step 4: Select a-stars and update the database.} A-stars are created by merging pairs in candidates that give the maximal gain. Then, the code table $CT_L$ and inverted database are updated. Then, {Steps 3 and 4 are repeated until there is no candidate leafset pair that can compress the database more.} {\color{black}A-stars left in model $M$ are the set of patterns we want, where an a-star with shorter code length will have a higher ranking.}

\begin{algorithm}[htbp]
\small
\LinesNumbered
    \SetKwFunction{Find}{Find}
    \SetKwInOut{Input}{input}\SetKwInOut{Output}{output}
    \Input{An attributed graph G with two parts: the adjacency list and the mapping function}
    \Output{Compressing a-star patterns}
    Encode all coresets $S_c^M$; \tcp{\textbf{Step 1}}
    
    $CT_c \leftarrow (S_c,Code_c)$\;
    $CT_L \leftarrow (S_L, S_c, Code_L)$\;
    
    $I \leftarrow (S_L, S_c, Position)$; \tcp{\textbf{Step 2}}
    
    \BlankLine
    $candidates \leftarrow$ generate\_candidates($S_L^M$);\tcp{Algorithm~\ref{alg:candidate}} 
    \Repeat{$candidates  =  \emptyset$}{
        \tcp{select pair with the max. gain $\Delta L$}
        $p \leftarrow candidates.pop()$\; 
        $I,CT_L \leftarrow$ merge($p$); \tcp{\textbf{Step 4}(update $I$ and $CT_L$)}
        $candidates \leftarrow$ generate\_candidates($S_L^M$); \tcp{\textbf{Step 3}}
    }
\Return{all a-stars in $M$}
\caption{The CSPM algorithm}
\label{alg:basic}
\end{algorithm}

\vspace{-6pt}

\begin{algorithm}[htbp]
\small
\LinesNumbered
    \SetKwFunction{Find}{Find}
    \SetKwInOut{Input}{input}\SetKwInOut{Output}{output}
    \Input{All leafsets $S_L^M$ in $CT_L$, inverted database $I$}
    \Output{A candidate list of leafset pairs}

    \BlankLine
    $candidates = []$\;
    $P \leftarrow$ enumerate($S_L^{M}$, 2)\;
    \For{$p \in P$}{
        $gain \leftarrow$ calculate\_gain($p$, $I$);   \tcp{Equation~\eqref{eq:gain}}
        \lIf{gain $>$ 0}{candidates.append($p$)}
    }
    Sort $candidates$ by descending order of gain\;
    \Return{$candidates$}
\caption{Generate leafsets merge candidates}
\label{alg:candidate}
\end{algorithm}

\section{Optimization}
\label{sec:optimization}

CSPM can mine a-star patterns efficiently. However, we found in empirical experiments that  attributes are many while  relationships between attribute values are sparse. Hence, few pairs can improve compression, among a huge number of possibilities.
 Besides, few gains are affected by a merge operation and need to be re-calculated and updated.
 Thus, it would be time-consuming to calculate all gains of all possible leafset pairs during each iteration.
 To improve the performance of the candidate generation process, an optimized CSPM version named CSPM-Partial is put forth. It updates only some of the gains and leafset pairs in candidates after each merge operation on account of the observation that two leafsets without a common coreset
 
 will never be merged even if they are merged as a new pattern later. More precisely, leafset pairs that are not adjacent to the same coreset will never have positive gains. Thus, only the gains of pairs with \emph{related} leafsets should be updated after each merge, instead of calculating the gain $C_n^2$ times, where $n$ is the leafset count in the inverted database. A leafset is called \emph{related} to a merged pair if it contributes a positive gain when merged with a leafset in the merged pair $p$. For this optimization, a dictionary $rdict$ is built to store all \emph{related} leafsets for each leafset of the inverted database. 
 
 Algorithm~\ref{alg:partial} shows the CSPM-Partial procedure. Line 6 creates the  $rdict$ dictionary, which stores $key,value$ pairs. Each key is a leafset and the value is the \emph{related} leafsets that can be merged with it in $candidates$.
 Compared with CSPM-Basic, the generate\_candidates function (line 9 in Algorithm~\ref{alg:basic}) is substituted by an update operation (line 10 in Algorithm~\ref{alg:partial}).

That way, after doing the merge operation in each iteration, only some gains are calculated and updated based on the previous $candidates$, instead of enumerating all possible pairs and re-calculating their gains again to generate a new $candidates$.

\begin{algorithm}[htbp]
\small
\LinesNumbered
    \SetKwFunction{Find}{Find}
    \SetKwInOut{Input}{input}\SetKwInOut{Output}{output}
    \Input{An attributed graph G with two parts: the adjacency list and the mapping function}
    \Output{Compressing a-star patterns}
    \BlankLine
    Encode all coresets $S_c^M$; 
    
    $CT_c \leftarrow (S_c,Code_c)$\;
    
    $CT_L \leftarrow (S_L, S_c, Code_L)$\;
    
    $I \leftarrow (S_L, S_c, Position)$; 
    
    \BlankLine
    $candidates \leftarrow$ generate\_candidates($S_L^M$);\tcp{Algorithm~\ref{alg:candidate}} 
    
    $rdict\leftarrow$ related\_dict($candidates$)
    
    \Repeat{$candidates  =  \emptyset$ or len($rdict$) $=0$}{
        $p \leftarrow candidates.pop()$\; 
        
        $I,CT_L \leftarrow$ merge($p$)\;
        
        $candidates, rdict \leftarrow$ update($p$, $I$); \tcp{Algorithm \ref{alg:update}}
    }
\Return{all a-stars in $M$}
\caption{The CSPM-Partial algorithm}
\label{alg:partial}
\end{algorithm}

Details about the update operation on $rdict$ and $candidates$ (after the pair $p$ is merged) are shown in Algorithm~\ref{alg:update}. 
$l_{total}$ and $l_{part}$ in line 1 contain the totally merged and partly merged leafsets of $p$, respectively. 
The overall update procedure consists of three operations (Remove, Add and Update).
First, totally merged leafsets and related pairs are removed from $rdict$ and $candidates$ (line 4). 
Next, all possible \emph{related} leafests ($rel$) are obtained  by intersecting $rdict[idx]$ and $rdict[idy]$ (line 6) based on the aforementioned observation. 
Each of them ($rel$) could be merged with the new leafset $new_a$. After calculating the gains, pairs with positive gains are added to $candidates$ and the \emph{related} information is recorded in $rdict$ (line 10-12). 
Finally, the pairs whose gains are influenced by the merge operation are updated in $candidates$ (line 17-21). It is a fact that frequencies of a-stars having partly merged leafsets are always reduced by the merge operation. 
Thus, the influenced pairs may not contribute to compression anymore, i,e. the gains are no longer larger than zero. For this reason, the corresponding pairs and their \emph{related} leafsets are removed from $candidates$ and $rdict$, respectively.

\begin{algorithm}[htbp]
\small
\LinesNumbered
    \SetKwInOut{Input}{input}\SetKwInOut{Output}{output}
    \Input{The merged pair $p$, the inverted database $I$}
    \Output{Updated $candidates$ and $rdict$}

    \BlankLine
    $l_{total}, l_{part} \leftarrow $ merge\_state($p$)\; 
    $idx, idy, new_a \leftarrow p[0], p[1], \{p[0] \cup p[1] \}$\; 
    
    \BlankLine
    \tcp{(1) \textbf{Remove} totally merged leafsets}
    \lIf{$l_{total} \neq \emptyset$}{
        delete $l_{total}$ in $candidates$, $rdict$}
    
    \BlankLine
    \tcp{(2) \textbf{Add} pairs with new leafset}
    \For{$rel \in rdict[idx] \cap rdict[idy]$}{
        $pair \leftarrow (rel, new_a)$\;
        $gain \leftarrow $ calculate\_gain($pair$, $I$) \tcp{Equation~\eqref{eq:gain}}
        \If{$gain>0$}{
            $rdict[new_a].add(rel)$\;
            $rdict[rel].add(new_a)$\;
            $candidates.add(pair)$\;
        }
    }
    
    \BlankLine
    \tcp{(3) \textbf{Update} influenced pairs}
    \If{$l_{part} \neq \emptyset$}{
        \For{$rel \in rdict[l_{part}]$}{
            $pair \leftarrow (rel, l_{part})$\;
            $gain \leftarrow $ calculate\_gain($pair$, $I$) \tcp{Equation~\eqref{eq:gain}}
            \leIf{$gain>0$}{
                update $pair.gain$ in $candidates$}{
                delete $pair$ in $candidates$, $rdict$}
        }
    }

    \Return{$candidates, mdict$}
\caption{Update operation}
\label{alg:update}
\end{algorithm}

\color{black}
We here analyze the complexity of CSPM-Basic and the optimized CSPM-Partial.  Given an attributed graph $G$ with $|E|$ edges, $|V|$ vertices, $|A|$ distinct attribute values. We denote $|\Bar{A}|$  the average number of values per attribute, $|S_L^M|$  the number of leafsets after merging, and $|F|$ the number of generated a-stars. The time and  space complexity of the two CSPM versions are analyzed in  as below.

\textbf{Time Complexity.} 
First, CSPM encodes all coresets. It takes $O(1)$ time to insert elements in $CT_c$.
Next, all the attributes are scanned in all vertices to construct the inverted database, which takes at most $O(|E| \times |\Bar{A}|^2)$. 
Then, patterns are generated by doing merges, using at most $(|S_L^M| - |A|)$ iterations. 
For each iteration, CSPM-basic generates candidates (Algorithm~\ref{alg:candidate}) in at most $O(|S_L^M|^2)$ steps (the last iteration). 
The merge operation is $O(1)$ (add/remove line). 
CSPM-Partial optimizes this procedure by only considering some candidate pairs,

and the worst case is $O(|S_L^M|)$ in each iteration. 
But generally, few attributes are highly correlated with the merged leaves. 
Thus, the time complexity of recalculating gains in each iteration is $O(k), k \ll |S_L^M|$. 
This is why CSPM-Partial outperforms CSPM-Basic in experiments.

\textbf{Space Complexity.} The dataset is stored as an inverted database, which is updated to generate a-stars. 
Thus, CSPM-Basic uses $O(|F|)$ space. The space complexity of CSPM-Partial is $O(|F| + |R|)$, where $|R|$ is the additional memory for keeping potentially related leaf-values, with the worst case of $O(|A|\times|A-1|)$.

But for real applications, it is unlikely that all attributes are correlated with each other due to sparsity. 
Thus, the additional memory usage is generally very small.

\color{black}

\section{Experiments}
\label{sec:experiments}

To evaluate the performance of the proposed CSPM algorithm {\color{black} Subsection~\ref{Exp:optimization}} first presents a runtime and  gain update ratio analysis of the basic CSPM (SPM-Basic), and the optimized version (CSPM-Partial).
\color{black}
Then, Subsection~\ref{exp:pattern} describes some simple a-stars found in real data to intuitively show that they are meaningful. The quality and usefulness of the a-stars as a whole are then evaluated in two application scenarios (Subsection~\ref{exp:completion} and \ref{exp:alarm}).

First, CSPM is quantitatively assessed for the popular graph attribute completion task.
Then, 
\color{black}
CSPM is used for an industrial application to extract alarm correlation rules for fault management in telecommunication networks. Note that, {\color{black}CSPM-Partial is adopted  for the two applications owing to its efficiency.} 

\subsection{Evaluation of the CSPM-Partial Optimization}
\label{Exp:optimization}
The first experiments were done to evaluate the benefits of using the proposed partial updating optimization.

\textbf{Algorithms.} Three algorithms were compared. The baseline algorithm is \emph{CSPM-Basic} without optimization.
The second algorithm is 
\emph{CSPM-Partial} (CSPM with the partial updating optimization).
Moreover, to provide a point of reference for experiments, the SLIM algorithm~\cite{smets2012slim:} is also adopted for comparison. In fact, although SLIM finds simple patterns (sets of values that co-occur) without considering inner relationship, it is close to this study as SLIM also is a compression-based algorithm and it can {\color{black}be easily applied to an attributed graph by treating} coresets in each adjacency list tuple as items.

To our best knowledge, a-star patterns have not been considered before. Note that, graph pattern mining algorithms and summarization techniques mentioned in Section~\ref{sec:related work} such as VOG and GraphMDL, focus on mining different patterns with complicated structure {\color{black}(shown in Table~\ref{tab:comparison})}, which {\color{black}is quite different from our work} and leads them to have a lower efficiency. Hence, these techniques are not included as baselines in this work.

\textbf{Datasets.} Four benchmark datasets having various characteristics were used. The statistics are summarized in  Table.~\ref{tab:statistics}, where $|S_c^M|$ gives the number of coresets 
in the inverted database. \emph{DBLP}~\cite{desmier2012cohesive} is a citation network indicating co-author relationships (edges) between researchers (vertices) during years 2006-2010. Attribute values are the conferences/journals a person has published in. The \emph{DBLP-Trend}~\cite{fournier2019mining} is a variant of \emph{DBLP} where the attribute indicates trends about publications. For example, (ICDE+, ICDE-, ICDE=) indicates that the number of publications in ICDE has increased, decreased, and stayed the same since the previous year, respectively. The \emph{USFlight}~\cite{kaytoue2014triggering} dataset contains data about flights (edges) between US airports (vertices) in 2005. The \emph{Pokec} dataset\footnote{https://stanford.io/3oZH9EI} indicates friendship relationships (edges) between persons (vertices) on the Pokec social network. The attribute values store the musical tastes of users.

\begin{table}[htbp]
\footnotesize
\centering
\caption{Statistics about datasets}
\label{tab:statistics}
\begin{tabular}{lcccc}
\toprule
Dataset       & DBLP     & DBLP-Trend & USFlight & Pokec      \\ \midrule
\#Nodes       & 2,723    & 2,723      & 280      & 1,632,803  \\
\#Total edges & 3,464    & 3,464      & 4,030    & 30,622,564 \\
$|S_c^M|$     & 127      & 271        & 70       & 914        \\
Category      & Citation & Citation   & Airport  & Music      \\ \bottomrule
\end{tabular}
\end{table}

\textbf{(1) \emph{Runtime}.} In the first experiment, the influence of the proposed optimization on the runtime was evaluated by comparing CSPM-Partial with CSPM-Basic. 
Results are shown in TABLE~\ref{tab:runtime}. 
Note that, the result of CSPM-Basic on the Pokec dataset is written as ($-$) because CSPM-Basic was not able to terminate after 48 hours. 
Three observations are made.

\begin{table}[htbp]
\footnotesize
\centering
\caption{Runtime comparison}
\label{tab:runtime}
\begin{tabular}{lccc}
\toprule
    & \multicolumn{3}{c}{Runtime (s)}      \\
Dataset           & SLIM      & CSPM-Basic & CSPM-Partial \\ \midrule
DBLP       & 4.69      & 43.13      & \textbf{0.98}         \\
DBLP-Trend & 48.69     & 956.61     & \textbf{25.46}        \\
USFlight   & 1.25      & 10.16      & \textbf{1.43}         \\
Pokec      & 166,678.3 & --         & \textbf{1,403.21}     \\ \bottomrule
\end{tabular}
\end{table}

First, the CSPM-Basic algorithm is approximately 10 times slower than SLIM. This result is reasonable since CSPM-Basic solves a problem that has two variables (core and leaf values) and it considers the graph topology which is more complex. Moreover, CSPM-Basic merges and updates candidate pairs without using any optimization.

\begin{figure}[htbp]
\vspace{-4pt}
\centering
\tiny
\includegraphics[width=0.50\textwidth]{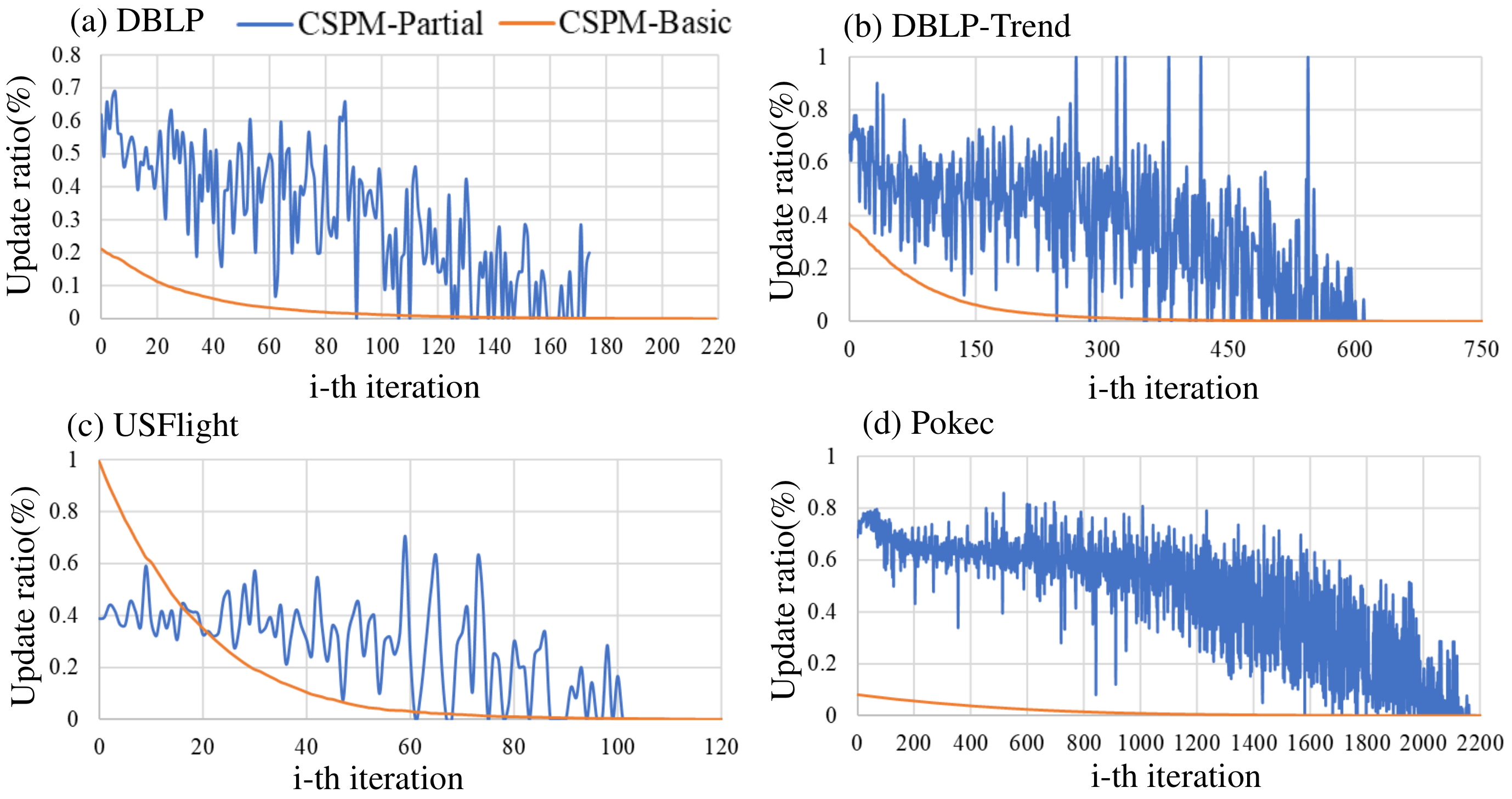}
\caption{Comparison of gain update ratio}
\label{fig:updateRatio}
\vspace{-2pt}
\end{figure}

Second, it was found that, the optimized algorithm, CSPM-Partial, is considerably faster than the CSPM-Basic. Especially for Pokec which is a large dataset. This result demonstrates that the proposed optimizations can make the algorithm more efficient.

A third observation is that, the runtime of CSPM-Partial increases as the number of coresets increases. 
In fact, CSPM-Partial optimizes the performance by generating several new leaf values during each iteration. However, the more leaf values that $CT_L$ contains, the more the frequency of gain pairs calculations for each iteration. Therefore, the runtime is increased. 

Overall, the optimized algorithm CSPM-Partial achieves high performance in terms of runtime especially for large datasets.

\textbf{(2) \emph{Gain update ratio}.} The second experiment is performed to compare the relative amount of gain calculations achieved by the two CSPM versions. Fig.~\ref{fig:updateRatio} shows the gain update ratio of the two algorithms for each dataset from the first to the last iteration. 
Recall that, during an iteration and after merging two leafsets, merge candidates are updated. Moreover, the gain update ratio represents the ratio of gain values that are added or updated out of the total number of possible calculations in each iteration. 

It can be observed that, CSPM-Partial often obtains a better gain update ratio compared with CSPM-Basic. This is why CSPM-Partial performs well on the used datasets such as DBLP and Pokec. 

\subsection{Pattern Analysis}
\label{exp:pattern}
To evaluate the interestingness of patterns discovered by the proposed CSPM algorithm, a manual inspection of the discovered patterns was performed. Besides, patterns with small code lengths are more frequent and have the strongest relationship between their core and leaf values.

\begin{figure}[htbp]
\centering
\includegraphics[width=0.49\textwidth]{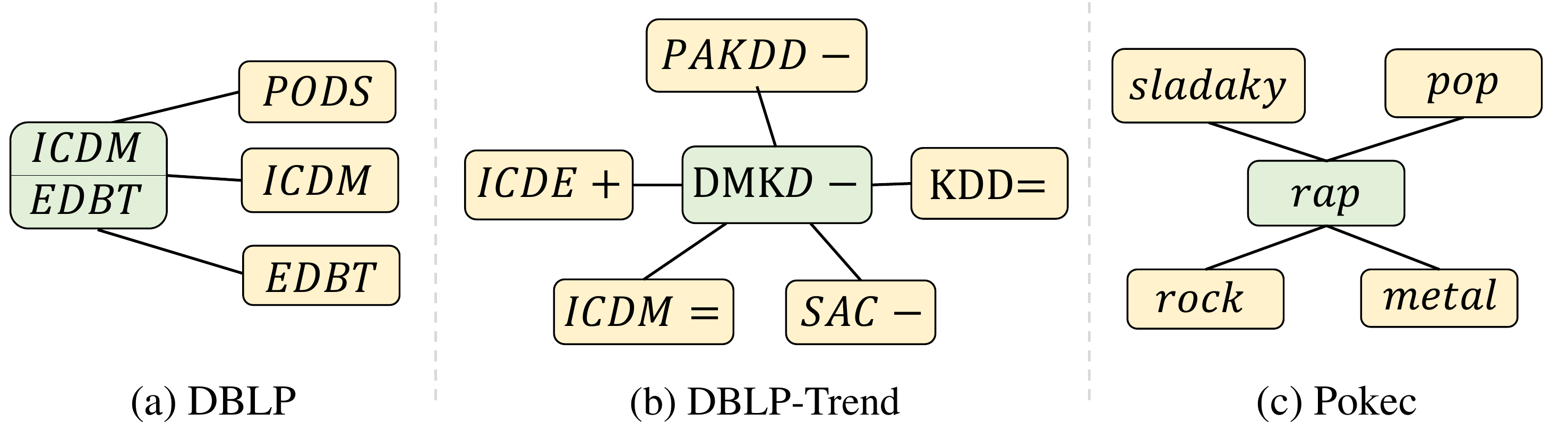}
\caption{{\color{black}Example a-stars from DBLP, DBLP-Trend and Pokec}}
\label{fig:patternAnalysis}
\end{figure}

\textbf{(1) \emph{DBLP}.}
Two patterns mined in DBLP and DBLP-Trend are depicted in Fig.~\ref{fig:patternAnalysis}(a). It was found that, there are many pattern occurrences where a researcher published a paper in $ICDM$ and $EDBT$ during a year, and his/her co-authors published in $PODS$, $ICDM$ and $EDBT$ during the same year. This is reasonable as $PODS$, $ICDM$, and $EDBT$  belong to the same research area (data-mining), and generally, co-authors of a paper are interested in the same research areas. 
\color{black}
The pattern of Fig.~\ref{fig:patternAnalysis}(b) from  DBLP-Trend can be interpreted in a similar way. But a difference is that patterns found in DBLP-Trend contain information about trends.
\color{black}

\textbf{(2) \emph{USFlight}.} An example pattern from the USFlight dataset is: $(\{NbDepart-\},$ $\{NbDepart+,$ $ DelayArriv-\})$, which means that, if the number of departure flights is reduced in an airport, there is a high chance that some connected airports will have an increase in their number of departure flights and a decrease in their number of delayed flights.

\textbf{(3) \emph{Pokec}.} Pokec dataset contains the music preferences of people from different communities. Some interesting patterns were also discovered in this dataset using CSPM such as: $(\{rap\}, $ $\{rock,$ $ metal,$ $ pop, $ $sladaky\})$. This pattern has a short encoding length and a high frequency which means that it greatly contributes to the reduction of the conditional entropy. This pattern seems reasonable because some music types such as rock, metal, pop, sladaky, and rap are preferred by young people. Another pattern is found and it shows the music preferences of older people: $(\{disko\},$ $\{oldies,$ $ disko\} )$. 

\subsection{Node Attribute Completion}
\label{exp:completion}
In this section, we quantitatively assess the patterns to check whether CSPM can reveal underlying characteristics of the data. More precisely, we apply CSPM to a graph-related classification task, namely node attribute completion~\cite{chen2020learning}. Attribute-missing graph completion is useful for numerous real-world applications. Thus, various algorithms have been proposed~\cite{chen2020learning,monti2017geometric,berg2017graph,csimcsek2008navigating} to deal with ths task. Intuitively, if attribute-stars extracted by CSPM can summarize well the data characteristics, incorporating mined patterns into the existing algorithms should improve or at least not degrade their performance.

We first present a scoring scheme which enables CSPM for the node completion task. It is based on the idea that a missing attribute value can be treated as a coreset with a core value and those of the neighbor nodes as leaf-values in an a-star. Then, a core value in an a-star with a smaller code length is more likely to be an attribute value of the targeted node. Therefore, we can score all possible attribute values using the code tables produced by CSPM. 

The scoring module is summarized in Algorithm~\ref{alg:score}. In brief, given the a-star set $M$ of CSPM and the attributes of the neighbors, the algorithm will return the scores of all possible attribute values for a targeted node $v$. In general, it is difficult to perfectly match the leaf-values in the leafset of an a-star with the attribute values of a neighbor. Thus, a weight is introduced to evaluate the similarity between leaf values in the leafset of an a-star and that of the neighbors of $v$. Intuitively, a leafset that is not similar to a set of neighbors will have a large weight $w$. As a result, it has a small score $cl$ which indicates that the corresponding core-values are less likely to be an attribute value of $v$. 

\begin{algorithm}[htbp]
\small
\LinesNumbered
    \SetKwFunction{Find}{Find}
    \SetKwInOut{Input}{input}\SetKwInOut{Output}{output}
    \Input{An attributed graph $G$ with  $n$ attribute values, the model $M$,  a node $v$ with missing attribute values}
    \Output{Scores for all possible attribute values}
    \BlankLine
    
    $scores \leftarrow [-\infty]_{n \times 1}$;  \tcp{Set scores to -$\infty$}
    
    $neighbors \leftarrow$ neighbor\_attributes($v$)\;
        
    \ForEach{a-star $S$ in $M$}{ 
        $S_{code} \leftarrow $ code\_length($S$); \tcp{Equation~\eqref{eq:code length}}
            
        $w \leftarrow $ similarity($S_L$, $neighbors$); \tcp{$S_L$:leafset of $S$}
        $cl \leftarrow - w \times S_{code}$\;
        \If{$cl >  scores[S_c]$}{ 
            $scores[S_c] \leftarrow cl$; \tcp{$S_c$:coreset of $S$}
            }
    }   
    \Return{$scores$}
    \caption{The scoring module of CSPM}
 \label{alg:score}
\end{algorithm}
\vspace{-5pt}

\begin{figure}[htbp]
\centering
\includegraphics[width=0.5\textwidth]{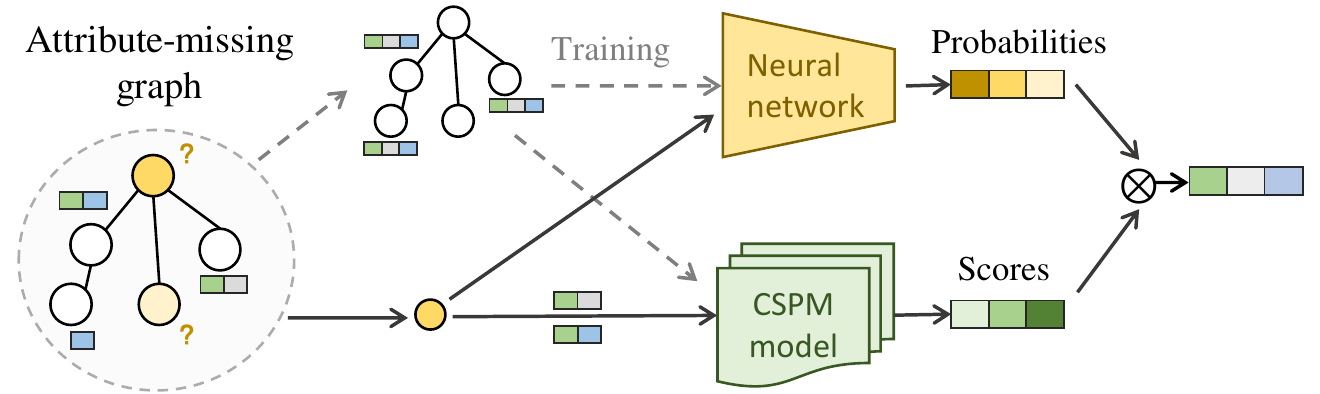}
\caption{The whole process of the attribute completion task using CSPM}
\label{fig:Attribute_completion}
\end{figure}

Based on the fact that CSPM is designed to learn the underlying correlations between attribute values, it is suitable to assist other attribute completion models. This is done by integrating CSPM with attribute scoring into existing node attribute completion models. The prediction process is illustrated in  Fig.~\ref{fig:Attribute_completion}.  
Besides, the output of a graph completion model for each vertex is a vector containing values indicating the possibilities that the attribute values appear while the vector obtained by the CSPM model records the scores of each attribute. In the proposed approach, the two vectors are normalized separately and then multiplied to get final scores.

Experiments are conducted on three benchmark citation networks in which attributes are categorical, namely  Cora \cite{2009Social}, Citeseer \cite{2008Collective}, and DBLP \cite{desmier2012cohesive}. Similarly to~\cite{chen2020learning}, the performance is evaluated according to the Recall@K and NDCG@K~\cite{0A}, where larger values indicate better results.
\color{black}
Specifically, the Recall reflects the correctness of the predicted attributes and the NDCG assesses ranking quality.
\color{black}

\color{black}

Note that DBLP is evaluated with a smaller K as it has less attribute values per node. The results are summarized in TABLE~\ref{tab:attributeCompletion}. It can be seen that, by integrating the a-stars-based scoring module into the attribute completion process, all the baseline algorithms are improved with different degrees. The most significant improvement is for NeighAggre or VAE which have relatively poor performance. These observations confirm that CSPM can successfully summarize the underlying characteristics of the given data.
\begin{table*}[htbp]
\centering
\caption{Profiling evaluation for node attribute completion}
\tiny
\label{tab:attributeCompletion}
\resizebox{0.79\textwidth}{!}{%
\begin{tabular}{|c|c|cccccc|}
\hline
\multicolumn{1}{|l|}{}     & \diagbox{Method}{Metric}                                  & Recall@10       & Recall@20       & Recall@50       & NDCG@10         & NDCG@20         & NDCG@50         \\ \hline
\multirow{13}{*}{Cora}     & NeighAggre~\cite{csimcsek2008navigating} & 0.0895          & 0.1396          & 0.1944          & 0.1203          & 0.1534          & 0.1832          \\
                           & CSPM+NeighAggre                          & 0.1339          & 0.1842          & 0.3002          & 0.1856          & 0.2190          & 0.2796          \\ \cline{2-8} 
                           & VAE~\cite{kingma2013auto}                & 0.0875          & 0.1215          & 0.2069          & 0.1219          & 0.1451          & 0.1901          \\
                           & CSPM+VAE                                 & 0.1023          & 0.1512          & 0.2471          & 0.1429          & 0.1758          & 0.2266          \\ \cline{2-8} 
                           & GCN~\cite{kipf2016semi}                  & 0.1261          & 0.1782          & 0.2930          & 0.1727          & 0.2075          & 0.2680          \\
                           & CSPM+GCN                                 & 0.1272          & 0.1843          & 0.3029          & 0.1749          & 0.2131          & 0.2754          \\ \cline{2-8} 
                           & GAT~\cite{velivckovic2017graph}          & 0.1281          & 0.1811          & 0.2982          & 0.1747          & 0.2108          & 0.2722          \\
                           & CSPM+GAT                                 & 0.1302          & 0.1862          & 0.3070          & 0.1780          & 0.2161          & 0.2794          \\ \cline{2-8} 
                           & GraphSage~\cite{hamilton2017inductive}   & 0.1224          & 0.1711          & 0.2790          & 0.1689          & 0.2018          & 0.2586          \\
                           & CSPM+GraphSage                           & 0.1264          & 0.1769          & 0.2918          & 0.1738          & 0.2075          & 0.2680          \\ \cline{2-8} 
                           & SAT~\cite{chen2020learning}              & 0.1461          & 0.2132          & 0.3375          & 0.2069          & 0.2514          & 0.3177          \\
                           & CSPM+SAT                                 & 0.1539          & 0.2184          & 0.3465          & 0.2131          & 0.2569          & 0.3244          \\ \cline{2-8} 
                           & Avg.improvement(\%)                       & \textbf{+12.94} & \textbf{+11.41} & \textbf{+14.57} & \textbf{+13.43} & \textbf{+12.56} & \textbf{+13.83} \\ \hline
\multirow{13}{*}{Citeseer} & NeighAggre~\cite{csimcsek2008navigating} & 0.0410          & 0.0737          & 0.1242          & 0.0658          & 0.0932          & 0.1274          \\
                           & CSPM+NeighAggre                          & 0.0574          & 0.0936          & 0.1742          & 0.0952          & 0.1255          & 0.1784          \\ \cline{2-8} 
                           & VAE~\cite{kingma2013auto}                & 0.0369          & 0.0662          & 0.1292          & 0.0580          & 0.0825          & 0.1288          \\
                           & CSPM+VAE                                 & 0.0438          & 0.0766          & 0.1480          & 0.0711          & 0.0985          & 0.1452          \\ \cline{2-8} 
                           & GCN~\cite{kipf2016semi}                  & 0.0613          & 0.1060          & 0.2000          & 0.1000          & 0.1370          & 0.1990          \\
                           & CSPM+GCN                                 & 0.0628          & 0.1076          & 0.2039          & 0.1027          & 0.1402          & 0.2031          \\ \cline{2-8} 
                           & GAT~\cite{velivckovic2017graph}          & 0.0535          & 0.0988          & 0.1958          & 0.0860          & 0.1238          & 0.1871          \\
                           & CSPM+GAT                                 & 0.0585          & 0.1039          & 0.1979          & 0.0943          & 0.1322          & 0.1935          \\ \cline{2-8} 
                           & GraphSage~\cite{hamilton2017inductive}   & 0.0561          & 0.1008          & 0.1939          & 0.0879          & 0.1250          & 0.1858          \\
                           & CSPM+GraphSage                           & 0.0579          & 0.1038          & 0.1971          & 0.0915          & 0.1297          & 0.1905          \\ \cline{2-8} 
                           & SAT~\cite{chen2020learning}              & 0.0657          & 0.1125          & 0.2103          & 0.1110          & 0.1501          & 0.2142          \\
                           & CSPM+SAT                                 & 0.0684          & 0.1164          & 0.2141          & 0.1151          & 0.1552          & 0.2192          \\ \cline{2-8} 
                           & Avg.improvement(\%)                       & \textbf{+12.97} & \textbf{+9.30}  & \textbf{+10.21} & \textbf{+14.57} & \textbf{+11.72} & \textbf{+10.52} \\ \hline
\multicolumn{1}{|l|}{}     & \diagbox{Method}{Metric}                                  & Recall@3        & Recall@5        & Recall@10       & NDCG@3          & NDCG@5          & NDCG@10         \\ \hline
\multirow{13}{*}{DBLP}     & NeighAggre~\cite{csimcsek2008navigating} & 0.2022          & 0.2648          & 0.3755          & 0.2583          & 0.3000          & 0.3631          \\
                           & CSPM+NeighAggre                          & 0.3685          & 0.4646          & 0.6028          & 0.4302          & 0.4963          & 0.5709          \\ \cline{2-8} 
                           & VAE~\cite{kingma2013auto}                & 0.2511          & 0.3119          & 0.4728          & 0.3071          & 0.3494          & 0.4370           \\
                           & CSPM+VAE                                 & 0.3066          & 0.3801          & 0.5276          & 0.3755          & 0.4266          & 0.5050          \\ \cline{2-8} 
                           & GCN~\cite{kipf2016semi}                  & 0.3994          & 0.4832          & 0.6657          & 0.4943          & 0.5523          & 0.6510           \\
                           & CSPM+GCN                                 & 0.5139          & 0.6400          & 0.7938          & 0.5916          & 0.6760           & 0.7582          \\ \cline{2-8} 
                           & GAT~\cite{velivckovic2017graph}          & 0.3964          & 0.4910          & 0.6574          & 0.4902          & 0.5539          & 0.6419          \\
                           & CSPM+GAT                                 & 0.5253          & 0.6634          & 0.8330           & 0.6018          & 0.6935          & 0.7814          \\ \cline{2-8} 
                           & GraphSage~\cite{hamilton2017inductive}   & 0.3445          & 0.4258          & 0.6069          & 0.4227          & 0.4793          & 0.5760          \\
                           & CSPM+GraphSage                           & 0.4020          & 0.4840          & 0.6460          & 0.4790          & 0.5345          & 0.6195          \\ \cline{2-8} 
                           & SAT~\cite{chen2020learning}              & 0.4890          & 0.6418          & 0.8227          & 0.5708          & 0.6694          & 0.7629          \\
                           & CSPM+SAT                                 & 0.4981          & 0.6445          & 0.8249          & 0.5780          & 0.6727          & 0.7664          \\ \cline{2-8} 
                           & Avg.improvement(\%)                       & \textbf{+30.68} & \textbf{+29.83} & \textbf{+20.80} & \textbf{+24.31} & \textbf{+24.52} & \textbf{+19.83} \\ \hline
\end{tabular}
}
\vspace{-5pt}
\end{table*}

\subsection{Alarm correlation analysis}
\label{exp:alarm}
Telecommunication networks are cornerstones of the communication systems in modern society. Ensuring high-quality services in complex and large telecom networks is important and challenging since millions of faults and alarms are triggered across the devices everyday. Thus, discovering relationships between these alarms and  filtering trivial alarms from important ones is critical to locate faults. 

Recently, several studies have applied pattern mining techniques for alarm correlation analysis. Wang et al.~\cite{wang2017efficient} designed and deployed an alarm management system called AABD (Automatic Alarm Behavior Discovery) where the alarm correlation module mines frequent patterns in alarm logs via sequential pattern mining~\cite{pei2004mining, truong2019fmaxclohusm}. The discovered patterns are then used to generate rules indicating that an alarm may be caused by another alarm. These rules are used to perform alarm compression (reduce the number of alarms presented to maintenance workers).
\color{black}
For instance, 
$($\texttt{Low\_signal}, $\{$\texttt{Link\_degrader}, \texttt{Microwave\_stripping}$\})$ is an alarm rule, suggesting that the cause alarm \texttt{Low\_signal} may be triggering derivative alarms (i.e., \texttt{Link\_degrader} and \texttt{Microwave\_stripping}). 
The alarm compression is achieved by only showing \texttt{Low\_signal} to the maintenance workers when they appear simultaneously. 
\color{black}

As AABD utilizes pattern mining algorithms which are not parameter-free, the quality of results is sensitive to the parameter values. 
Moreover, AABD ignores the connections between network devices and {\color{black}the alarm importance in a mined rule (i.e., cause alarm or not)} have to be inferred by the experts knowledge. To solve the above problems, the ACOR algorithm~\cite{fournier2020discovering} models alarm data as a dynamic attributed graph, and then extracts paired alarm having a high correlation by {\color{black}a tailored correlation measure} which also give the importance of each alarm in an alarm pair. 
\color{black}
Though ACOR can generate alarm rules by combining alarm pairs with the same cause alarm,  information loss is inevitable.
\color{black}

In this section, we evaluate the efficiency of CSPM for alarm correlation analysis by comparing with ACOR~\cite{fournier2020discovering}, checking whether the rules in the AABD rule library can be rediscovered with higher rankings. For this purpose, the coverage ratio~\cite{fournier2020discovering} is used as an evaluation metric. It is defined as follows: $coverage=\left|A\cap B\right|/\left|A\right|$, where $A$ is the set of valid rules and $B$ is the rules found by CSPM or ACOR. A high coverage ratio indicates that most of rules discovered by the proposed framework are valid. Hereafter, the coverage ratio for top-K represents the ratio of valid rules in the k rules that are found by CSPM or ACOR and have the highest correlation scores. Note that, the alarm patterns extracted by CSPM are star-shaped {\color{black}where the core serves as the cause alarms}, while the alarm patterns extracted by ACOR have a paired format. Hence, a-stars mined by CSPM are split into pairs.
\color{black}
Note that the results are not influenced because the rankings and scores of all alarm rules are maintained. It is only done to compare with ACOR, which mines pairwise alarm rules.
\color{black}

\begin{figure}[htb]
\vspace{-13pt}
\centering
\includegraphics[width=0.36\textwidth]{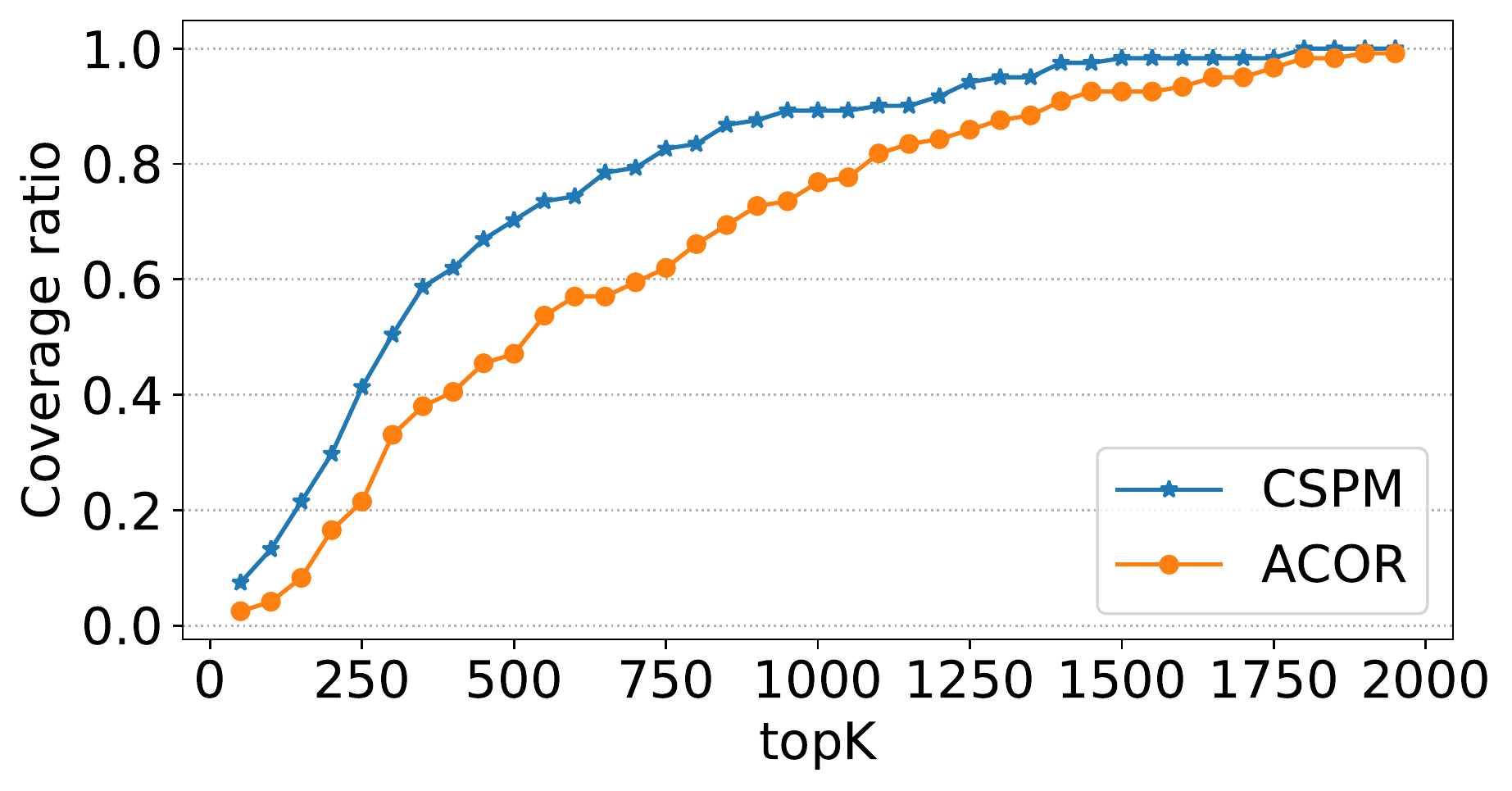}
\caption{The coverage ratio of ACOR and CSPM for alarm correlation analysis}
\label{fig:coverage_ratio}
\vspace{-3pt}
\end{figure}

The experiments were done on a real alarm dataset with 6 million triggered alarms collected from a metropolitan city of Southeast Asia from the 12$^{th}$ to 16$^{th}$ April, 2019. The collected alarms are categorized into 300 types, which fall into 11 rules stored in the AABD system~\cite{wang2017efficient}. These rules are then decomposed into 121 pair-rules~\cite{fournier2020discovering}. The coverage ratios of the two algorithms are given in Fig.~\ref{fig:coverage_ratio}. It can be found that, the coverage ratio increases as more rules are selected, and finally all the valid rules are found. Moreover, as it is expected, the valid rules found by CSPM are ranked higher comparing with the valid rules found by ACOR. This is due to the fact that ACOR evaluates each pair-rule separately. However, the proposed algorithm simultaneously and systematically takes all rules into account via the MDL.   

\section{Conclusion}
\label{sec:conclusion}

To find patterns that capture strong relationships between attributes in attributed graphs, this paper proposed a parameter-free and compression-based algorithm, named CSPM. It finds  star-shaped attribute patterns that best compress the data according to the conditional entropy and MDL principle. Experiments on large graphs have shown that CSPM can  identify interesting patterns and that its optimization improves its performance. Besides, CSPM was shown to boost the performance of state-of-the-art graph attribute completion models and find interesting patterns for alarm correlation analysis.

Some interesting possibilities for future work are: 
(1) to utilize a-stars found by CSPM for other graph-related learning problems such as graph classification, (2) to extend CSPM to handle other graph types such as dynamic attributed graphs and graphs with multiple attributes per edge.
(3) to develop a distributed version of CSPM to process very large databases.

\bibliographystyle{elsarticle-num}
\bibliography{reference.bib}


\end{document}